\documentclass[MS]{macro/neu_msthesis}

\usepackage{siunitx}
\usepackage{graphics} 
\usepackage{graphicx} 
\usepackage{epsfig} 
\usepackage{lipsum} 
\usepackage{amsmath, bm} 
\usepackage{amssymb}  
\usepackage{upgreek}
\usepackage{url}
\usepackage{caption}
\usepackage{subcaption}
\usepackage{pgfplots}
\usepackage{floatrow}
\usepackage{float}
\usepackage[utf8]{inputenc}
\usepackage{multicol}
\usepackage{tikz}
\usepackage{booktabs}
\usepackage{longtable}
\usepackage{algorithm}
\usepackage{algpseudocode}
\usetikzlibrary{shapes, arrows}
\usepackage{amsmath}
\newfloatcommand{capbtabbox}{table}[][\FBwidth]

\algrenewcommand\algorithmicwhile{\textbf{while}}
\algrenewcommand\algorithmicdo{\textbf{do}}
\algrenewcommand\algorithmicif{\textbf{if}}
\algrenewcommand\algorithmicthen{\textbf{then}}
\algrenewcommand\algorithmicelse{\textbf{else}}
\algrenewcommand\algorithmicend{\textbf{end}}

\title{Enabling Autonomous Navigation in a Snake Robot through Visual-Inertial Odometry and Closed-Loop Trajectory Tracking Control}

\author{Mohammed Irfan Ali}

\dept{Electrical and Computer Engineering}

\degree{Master of Science}

\degreename{Robotics}

\field{Robotics}

\submitdate{December 2025}

\numberofmembers{3}

\principaladviser{Dr. Alireza Ramezani}
\firstreader{Dr.xxx}

\chairman{Dr. xxx}

\dean{Dr. xxx}

\usepackage{amsfonts}			
\usepackage{times}		

\newcommand{\ifno}[1]{}

\usepackage{multirow}
\makeindex
\usepackage{makeidx}
\usepackage{acronym}

\usepackage{url}
\usepackage{graphicx}

\ifnum\pdfoutput>0
\usepackage[pdftex,
bookmarks=true,
bookmarksnumbered=true,
hypertexnames=false,
breaklinks=true           
]{hyperref}
\else
\usepackage[hypertex,
bookmarks=true,
bookmarksnumbered=true,
hypertexnames=false,
breaklinks=true           
]{hyperref}
\fi

\hypersetup{				
pdfauthor = {\authorRef},
pdftitle = {\titleRef},
pdfsubject = {\expandafter{\degreeRef} thesis submitted to Northeastern University},
pdfkeywords = {Multimodal robots,M4,Traversability,3D path planning}
pdfcreator = {LaTeX with hyperref package},
pdfproducer = {dvips + ps2pdf}}

\begin{document}

\pdfbookmark[1]{Cover}{cover}

\titlepage

\begin{frontmatter}

\pdfbookmark[1]{Table of Contents}{contents}
\tableofcontents
\listoffigures
\newpage\ssp
\listoftables



\begin{acknowledgements}

I would like to express my deepest gratitude to my advisor, Prof. Alireza Ramezani, for his mentorship, guidance, and confidence in my abilities throughout this thesis. His vision, technical insight, and continual support have shaped both this work and my growth as a researcher. I am also especially grateful to Adarsh Salagame, whose thoughtful mentorship and willingness to help at every stage have played a central role in the development of this thesis.

I am sincerely thankful to all the PhD mentors and members of the \textbf{SiliconSynapse Lab}, whose encouragement and insight have enriched this journey. I would especially like to thank Hrigved Suryawanshi, Henry Noyes, Bibek Gupta, and Kaushik Krishnamurthy for their consistent support, collaboration, and readiness to assist whenever needed.

I would also like to extend my appreciation to Prof. Hanumant Singh and Prof. Miriam Leeser for kindly agreeing to serve on my thesis committee, for their valuable feedback, and for the thoughtful discussions during my defense, which significantly strengthened this work.

I am grateful to my friends Haard Shah, Mihir Chaulkar, and Rayyan Rafikh for being a constant source of motivation, encouragement, and positivity throughout my master’s program.

Finally, I owe my deepest appreciation to my parents, uncle, aunt, and brother and my entire family. Their unwavering love, belief in me, and constant encouragement have been the foundation of everything I have accomplished. This work would not have been possible without their support.

\end{acknowledgements}


\begin{abstract}

Snake robots offer exceptional mobility across extreme terrain inaccessible to conventional rovers, yet their highly articulated bodies present fundamental challenges for autonomous navigation in environments lacking external tracking infrastructure. This thesis develops a complete autonomy pipeline for COBRA, an 11 degree-of-freedom modular snake robot designed for planetary exploration. While the robot's biologically inspired serpentine gaits achieve impressive mobility, prior work has relied entirely on open-loop teleoperation. This approach integrates onboard visual-inertial SLAM, reduced-order state estimation, and closed-loop trajectory tracking to enable autonomous waypoint navigation. A depth camera paired with edge computing performs real-time localization during dynamic locomotion, validated against motion-capture ground truth to characterize drift behavior and failure modes unique to snake robot platforms. A reduced-order framework estimates Center-of-Mass pose, driving a closed-loop controller that modulates CPG gait parameters through distance-dependent yaw error blending. Physical experiments validate the complete system, demonstrating accurate multi-waypoint tracking and establishing foundations for autonomous snake robot navigation.

\end{abstract}

\end{frontmatter}


\pagestyle{headings}



\chapter{Introduction}
\label{chap:introduction}


The exploration of extraterrestrial environments, such as lunar craters and icy terrains, presents unique mobility challenges that traditional wheeled rovers often cannot overcome. These environments contain steep slopes, loose regolith, unpredictable cavities, and regions where the terrain geometry is highly discontinuous. Ground robots designed with conventional rigid bodies or low-DOF mechanisms frequently become limited by their inability to distribute contact forces dynamically or adapt their morphology to changing surface conditions. As a result, robotics research has increasingly turned toward bio-inspired locomotion strategies that mimic organisms capable of thriving in such harsh environments.

Among the many biologically-inspired solutions, snake-like robots have emerged as a particularly promising class of systems \cite{8460584} \cite{9992457}. Their elongated, articulated bodies allow them to exploit redundancy, maintain multiple simultaneous contact points with the ground, and generate propulsion even in regions where the environment offers minimal footholds \cite{10694698} \cite{polybot} \cite{slithering_locomotion}. Through specialized gaits such as lateral undulation, concertina, sidewinding, and rectilinear locomotion, real snakes demonstrate extraordinary adaptability on sand, rocks, gravel, and narrow, cluttered passages \cite{Owen_1994}. Translating these capabilities into robotic platforms opens the door to new modes of planetary access and mobility, enabling robots to traverse surfaces and subsurfaces that are inaccessible to rovers or legged machines \cite{hyper_redundant_robot_locomotion}.

However, building a snake-like robot that can autonomously operate in extreme terrain is far from trivial \cite{10611384}. Most existing serpentine robots rely heavily on teleoperation, scripted gait execution, or external motion-capture systems to maintain stability and track motion. Their high number of degrees of freedom introduces challenges in perception, estimation, modeling, and control. Without the ability to localize themselves and understand their environment, these robots cannot fully exploit their mechanical capabilities. For planetary exploration missions—where communication delays, occluded GPS signals, and restricted bandwidth are unavoidable—true autonomy becomes a strict requirement. The robot must be able to map, localize, and plan onboard, without external infrastructure.

In this thesis, we focus on COBRA (Crater Observing Bio-inspired Rolling Articulator), an 11-degree-of-freedom (11-DOF) modular snake robot developed at Northeastern University \cite{cobraAISJ} \cite{10637166}. COBRA’s design draws on biological principles to enable locomotion across highly variable terrain. Its mechanical structure supports a range of gaits, including sidewinding which is useful for efficient traversal on granular or low-friction substrates, vertical undulation worm gaits for confined environments, and a unique tumbling mode enabled by magnetically linking the head and tail modules \cite{modulation_orthogonal_body} \cite{geometric_swimming}. In this configuration, the robot can behave like a passive rolling body, using gravity to descend steep slopes while conserving energy. These versatile modes make COBRA an attractive platform for crater exploration, sub-surface access, and inspection tasks where adaptability and robustness are critical.

Despite these strengths, COBRA currently lacks the autonomy required for deployment in environments where external tracking systems are unavailable. The robot’s locomotion has been successfully demonstrated in open-loop and teleoperated modes, but the absence of onboard perception limits its operational envelope. To close this gap, the robot must be equipped with sensing, state estimation, and feedback control components that together allow it to navigate without human intervention. For serpentine robots in particular, this represents a major challenge: their complex motion patterns often cause rapid perspective changes, large oscillations in sensor viewpoints, and self-occlusions, all of which complicate visual perception and introduce drift in estimation algorithms.

\section{Thesis Objectives}
\label{sec:thesis-objectives}

The objectives of this work are organized around enabling and evaluating autonomy on the COBRA platform. The primary goals are as follows:

\begin{itemize}
    \item Integrate a depth camera and edge computing hardware into the robot’s head module to establish a functional onboard visual-inertial odometry pipeline capable of real-time state estimation.
    \item Quantitatively evaluate the perception system against ground-truth motion capture data during diverse locomotion modes, including sidewinding, worm gaits, and survey maneuvers, to assess trajectory drift and scale consistency.
    \item Develop a reduced-order state estimation logic that processes head pose data to calculate the robot’s Center of Mass and define a dynamic bounding box for spatial planning.
    \item Formulate and validate a closed-loop control strategy that modulates CPG parameters in real time to actively steer the robot through a sequence of waypoints based on the estimated center of mass.
\end{itemize}

\section{Thesis Contributions}
\label{sec:thesis-contributions}

Building on the objectives outlined above, this thesis makes four primary contributions:

\begin{itemize}
    \item Demonstration of the feasibility of running computationally intensive 3D SLAM algorithms on compact edge hardware during the highly dynamic and oscillatory motion characteristic of snake robots.
    \item Identification of specific visual perceptual challenges unique to snake gaits, identifying specific failure modes and drift characteristics and providing a statistical breakdown of errors to establish a baseline for serpentine state estimation.
    \item Development of a simplified control architecture that abstracts complex multi-joint kinematics into a stable Center of Mass reference frame, enabling efficient trajectory tracking without full-body modeling.
    \item Experimental validation of a weighted waypoint-following strategy that dynamically balances relative bearing for path convergence and absolute orientation for final pose alignment to ensure precise autonomous navigation.
\end{itemize}

    
    
    

\chapter{Literature Review}
\label{chap:Literature Review}


For many years, planetary exploration has mainly depended on wheeled rovers like NASA’s Curiosity and Perseverance. These systems set strong mobility benchmarks and have produced valuable scientific data. However, they still face practical limits. Certain terrains, such as steep crater walls, loose granular soil, lava tubes, and permanently shadowed lunar polar regions, pose serious hazards. On these surfaces, rovers can sink into soft soil, get high-centered, or even tip over.

Shackleton Crater at the Moon’s south pole illustrates the challenge well. It is scientifically attractive for its possible ice deposits, yet the terrain is extremely treacherous. Regolith in these high-latitude craters behaves like a complex granular medium where  it can flow or collapse under pressure. Traditional models often fail to predict this behavior. The danger is that wheel or leg contact points can concentrate force on the soil, causing it to yield and risking mission failure. 

One promising direction is to take inspiration from biology. Limbless creatures like snakes distribute their weight and traction along a long, continuous body. This redundancy keeps any one section from applying too much pressure to the ground. As a result, a snake robot can squeeze through tight or cluttered spaces and support itself on many points simultaneously \cite{Liljeback2012SnakeRM}. 

To capture these benefits in a robot, design must be rethought. Instead of treating the body and controller as separate entities, a morpho-functional approach blends them together \cite{wang2025dynamicquadrupedalleggedaerial}. In this view, the robot’s physical structure itself helps to control the motion, an idea called embodied (or mechanical) intelligence. The COBRA robot (Crater Observing Bio-inspired Rolling Articulator) at the Silicon Synapse Laboratory exemplifies this concept. COBRA combines sidewinding (a snake-like gait) with a specialized tumbling mode for descending steep slopes. Its design can also use vision-in-the-loop control, where onboard cameras help adjust its movement in real time. 

Several of my lab’s other platforms highlight how morphology and control can develop together to support versatile mobility \cite{salagame2022letterprogresshuskycarbon}. For example, the M4 robot (Multi-Modal Mobility Morphobot) demonstrates “locomotion plasticity \cite{sihite2023demonstratingautonomous3dpath}\cite{sihite2022efficientpathplanningtracking}.” M4 can transform from rolling on the ground to flying in the air \cite{sihite2023m4}. It uses a nonlinear model predictive controller to manage each transition, handling different dynamics and actuator limits in each mode. This mode-switching is directly relevant to COBRA, which switches between an open-chain sidewinding configuration and a closed loop tumbling ring. The same motors and joints perform very different roles in these two modes, much like M4 repurposes its wheels as propellers. The M4 case underscores the lab’s central idea, which is that achieving autonomy in complex terrain requires tightly coupling morphology with predictive control. In other words, a robot’s body can be designed to simplify the control problem rather than complicate it.

Another example from the lab is Harpy, a robot that combines traditional legged walking with small thrusters on its torso \cite{11108068} \cite{10769831} \cite{10637139}. By firing these thrusters to generate aerodynamic forces, Harpy stabilizes its gait and resists disturbances that would topple a purely legged walker \cite{control_thruster_assisted} \cite{pitroda2024quadraticprogrammingoptimizationbioinspired}. Its control system relies on precise state estimation, momentum-based disturbance observers, and capture-point algorithms \cite{pitroda2024enhancedcapturepointcontrol}. These innovations offer lessons for COBRA. Just as Harpy uses thrust to compensate when its feet have poor traction, COBRA can change its body shape, lifting certain segments or adjusting its undulation amplitude to cope with slipping on weak regolith \cite{pitroda2024capturepointcontrolthrusterassisted}.

Husky Carbon, a quadrupedal robot from the lab, demonstrates another form of multi-modal locomotion by combining legs and thrusters \cite{salagame2022letterprogresshuskycarbon} \cite{krishnamurthy2024thrusterassistedinclinewalking} \cite{11107989}. Inspired by birds that flap their wings while running up steep hills, Husky uses small propellers mounted on its body to help it walk up slopes as steep as 45 degrees \cite{10644231} \cite{krishnamurthy2024enablingsteepslopewalking}. What makes Husky particularly interesting is that its legs can fold outward, repositioning the knee motors to act as aerial propellers. The robot essentially reconfigures itself from a walking quadruped into a flying quadrotor. This physical transformation between locomotion modes parallels COBRA's ability to switch from an extended sidewinding snake into a closed rolling ring. Both robots show that the same mechanical structure can be repurposed for entirely different gaits. For COBRA's operations in confined spaces and challenging terrain, Husky's approach offers an important lesson which is about when one locomotion mode encounters difficulty, such as legs slipping on steep slopes or a snake losing traction on loose surfaces, auxiliary actuation can maintain mobility. Husky relies on thrusters while COBRA uses dynamic body reshaping, but both avoid requiring fundamentally different hardware for each scenario \cite{10637015} \cite{krishnamurthy2024optimizationfreecontrolground}.

Northeastern's Aerobat brings the morpho-functional idea into the air \cite{9197376} \cite{biomimetic_robotic_platform} \cite{robotic_bat_flight} \cite{7353772} \cite{ramezani_2016_knwvh-d7f12}. It uses a bat-inspired wing with a clever mechanical linkage that produces complex flapping mostly passively, while small actuators steer the wing motion \cite{sihite2022wakebasedlocomotiongaitdesign} \cite{7487491}. This is a clear example of a “computational structure” where the mechanics of the robot perform much of the control automatically. This concept parallels COBRA’s design, where passive compliance, directional friction, and carefully shaped gaits handle much of the motion without extra computation. Aerobat also has “conjugate momentum observers” that infer wing interaction forces without direct force sensors. This suggests a strategy for COBRA’s sensing. On the Moon, a tactile skin might be fragile in a sharp regolith, so COBRA could rely on joint-torque sensors and observers to estimate terrain stiffness and slope from its own body forces. 

Leonardo, developed at Caltech, combines legged locomotion with onboard propellers \cite{kim2021bipedal}. It synchronizes high-frequency propeller thrusts with slower leg movements to perform dynamic tasks like slacklining and balancing. Its control system manages interactions across these different time scales and contact forces. This approach is relevant for COBRA’s design, where although the visual perception (e.g. SLAM processing) might run at only tens of hertz, the gait controller updates at hundreds of hertz. Leonardo shows that a unified control framework is needed to manage such disparate rates and actuation modes together. This insight is crucial for COBRA, which must integrate visual feedback into its high-rate locomotion controller. 

COBRA itself brings all these design themes together for the lunar environment. It is a modular, hyper redundant snake robot capable of alternating between sidewinding to traverse loose soil and a tumbling hexagonal-ring mode for descending steep crater walls \cite{10637168}. Its active joints and latching mechanisms allow it to reconfigure its shape on demand. Recent work has demonstrated vision-guided locomotion and manipulation on COBRA, as well as closed-loop control of its tumbling direction \cite{salagame2024validationtumblingrobotdynamics}. The next big leap is to bring vision fully into its sidewinding gait. In other words, enabling COBRA to use onboard cameras and inertial data to actively correct for slip and heading errors in real time is an open engineering challenge and the core focus of this research. 

Other snake-like robots highlight different design trade-offs. For example, NASA JPL’s EELS robot is designed for icy vents. It uses screw-driven propulsion and active skin elements to move on ice, along with a robust autonomy system for risk assessment. COBRA, by contrast, relies more on passive design: its skin materials and reconfigurable morphology provide traction and adaptation, rather than powered skin actuators. The field of soft snake robotics broadens the picture even further. Some soft snakes have fully deformable bodies or even “growing” capabilities, and many use kirigami-inspired skins. These skins have patterned cuts that pop into scale-like grips when flexed, greatly enhancing traction without additional motors. Although COBRA is currently a rigid-link robot, adding such smart materials could significantly improve its grip on loose soil. Future designs may even blend rigid and soft elements for maximum adaptability. 

Planning COBRA’s movement starts with the basics of snake locomotion. Unlike wheeled or legged robots, snakes generate motion through waves that travel along their bodies. The classical lateral undulation gait (often called the “serpenoid” model) produces a smooth, travelling wave of bending \cite{lateral_undulation}. In this gait, each joint follows commands that set the wave’s amplitude, frequency, and phase. On firm, high-friction ground, lateral undulation can effectively push the robot forward, since many body segments are in contact. However, on loose granular terrain, this method tends to fail as parts of the body simply slide or dig into the sand rather than advancing \cite{Marvi_2014}. 

Desert snakes avoid this problem by using sidewinding. Sidewinding coordinates a horizontal bending wave with a vertical lifting wave. In practice, this means the snake lifts some body segments off the ground in sequence, so that only certain patches touch at any given time. That contact pattern greatly reduces lateral 2 shear forces on the substrate and prevents it from collapsing. In effect, sidewinding lets the snake “roll” forward in segments, causing the sand to yield momentarily under one part of the body and then stabilize as the snake moves. 

In a robot, sidewinding is implemented by superimposing horizontal and vertical sinusoidal waves along the body. The waves are phased so that the parts of the body with the highest horizontal curvature are lifted by the vertical wave. The amplitude of the vertical wave is critical: if it is too small, the body will drag along the ground and disturb the substrate; if it is too large, the robot may become unstable and tip \cite{snake_gait_synthesis}. Simulation studies for COBRA show that the optimal wave parameters depend on factors like slope angle, soil compaction, and sinkage. In other words, a set of parameters that works on flat ground might fail on a slope. This means that static tuning is insufficient for real missions and COBRA will need to adapt its gait in real time to changing terrain. 

The robot’s surface materials also play a major role in locomotion. Biological snakes rely on frictional anisotropy: their scales grip well sideways but slide easily forward. Robotic snakes mimic this with directional skins or scale-like panels. A promising innovation is kirigami-inspired skin which is a flexible covering with patterned cuts that pop up into little hooks when the body bends. These hooks grip firmly when in contact, then flatten to reduce drag while sliding. In other words, this smart skin gives the robot passive mechanical intelligence, automatically increasing traction where needed without extra actuators. Another key design element is compliance in the actuation. By placing springs in series with the motors (series-elastic actuation), the robot gains flexibility. The springs absorb shocks during tumbling, protecting gears and motors, and can store and release energy with each wave cycle. From a control perspective, compliance makes force control simpler because the robot can physically conform to bumps in the terrain. Integrating these material and actuator features into COBRA would significantly enhance its robustness and efficiency on unpredictable lunar soil. 

Turning while sidewinding is another challenge. A simple approach which is shifting the wave phase to steer often causes the robot to roll or tip over on loose ground. Instead, the most reliable turning method is to modulate the wave amplitude along the body. This creates a cone-like contact pattern so the snake robot follows a smooth curved path. Other tricks, like changing the wave frequency or phase, can produce sharper turns but tend to destabilize motion on slippery soil. For COBRA’s missions, which require smooth and safe navigation, amplitude modulation is the most consistent way to correct its trajectory. 

Modeling how COBRA interacts with granular soil is essential for prediction and control. Simulating every grain of sand in detail would be too slow, so we use practical approximations like Granular Resistive Force Theory (RFT) \cite{9345981} \cite{rft_fluids} \cite{undulatory_swimming}. RFT divides the snake’s body into many small elements and sums up the empirical force on each element based on its orientation, depth, and velocity. Modern RFT models can handle three dimensional motions and even include cohesive forces, which matter because lunar regolith behaves differently than Earth sand (its particles are jagged and can stick together electrically) \cite{Li_2013}. RFT lets us predict failure modes—such as when the shear force on the ground exceeds the soil’s yield point and causes an avalanche. In that sense, RFT provides a simplified “plant model” that controllers or learned policies can use to anticipate slip and adapt the gait accordingly. 

A higher-level perspective comes from geometric mechanics, which treats locomotion as the mapping from internal shape changes to net motion \cite{geometric_swimming} \cite{Hatton2011GeometricMP} \cite{Kelly1995GeometricPA}. When the robot executes a cycle of joint motions (a gait), it traces a loop in its shape space that produces displacement. Researchers have developed tools to visualize how different loops yield different motions, highlighting which gaits give the most forward motion or the best 3 climbing efficiency under given constraints. For COBRA, using geometric mechanics means we can design its wave patterns based on the underlying physics rather than just by trial-and-error or pure biomimicry. In practice, these tools help us analytically find combinations of horizontal and vertical waves that best exploit the local soil dynamics \cite{Hatton2013GeometricVO} \cite{snakeboard}. 

Understanding how COBRA moves is only half the challenge; the robot also needs to know where it is. In GNSS-denied, dark, and dusty environments like lunar craters, reliable state estimation must come from onboard sensors. The standard solution is Visual-Inertial SLAM (VI-SLAM), which fuses camera images with high-rate IMU (inertial) data to estimate the robot’s pose \cite{biomimetics9110710} \cite{Qin_2018} \cite{keyframe_nonlinear_optimization} \cite{multi_state_kalman_filter} \cite{Campos_2021}. COBRA’s snake-like motion makes purely visual tracking difficult. As the head oscillates, camera images can blur or shift rapidly, breaking feature matching. The IMU helps bridge this gap by providing continuous motion data between frames, stabilizing the estimates. For a sidewinding snake that undergoes large rotational accelerations at its head, tightly coupled visual–inertial fusion is essential. 

In practice, the system must handle the different data rates of the sensors. IMUs sample very quickly, while cameras provide frames at a slower rate \cite{Forster_2017} \cite{Forster_2017}. The many high-rate inertial readings between frames are summarized into a compact constraint through inertial preintegration. This compresses thousands of IMU samples into a single motion estimate, keeping the SLAM computation efficient and stable. Among VI-SLAM methods, feature-based low-latency systems work well for the “inner loop” of odometry, giving fast pose updates to the controller. Meanwhile, denser mapping methods build a rich local map for navigation and obstacle avoidance. For COBRA, a hybrid architecture is ideal where it can use quick feature-based odometry for real time control, and maintain a denser local map for higher-level planning. 

The robot’s head motion also creates specific perception challenges. Rapid swings can produce motion blur or rolling-shutter distortion in the camera. The lighting in craters is often extremely high-contrast (bright sunlight next to deep shadow), which can confuse classical vision algorithms. The robot’s own body segments may even move into the camera’s view, causing self-occlusion. To address these issues, modern vision techniques are useful. Learned feature detectors and matchers (deep-learning-based) are more robust to blur and lighting changes than classical corner detectors. There are also “dynamic SLAM” methods that detect and mask out moving objects (including the robot’s own moving parts or sliding debris). In COBRA’s case, these would prevent the robot’s body or falling dust from corrupting the map. 

Another approach is to use an event camera instead of a conventional frame camera \cite{Gallego_2022} \cite{Mueggler_2017}. Event cameras report per-pixel brightness changes asynchronously with microsecond latency and very high dynamic range \cite{evo_event_camera}. They essentially avoid motion blur, since they capture changes rather than full images. An event-based visual inertial system could greatly improve tracking of the rapidly moving head. The tradeoff is that new algorithms and more processing power are needed to handle the event data, but this path shows great promise for agile perception. 

These perception challenges also affect control. A naive controller that tries to correct the raw camera pose (which jumps around with the gait) would end up “fighting” the robot’s natural oscillations and destabilizing it. The solution is to separate the fast rhythmic motion from the robot’s average movement. One approach is to define a “locomotion frame” that follows the robot’s mean pose over each gait cycle. By transforming pose estimates into this body-level frame, the rapid gait oscillations are filtered out. Control laws and guidance algorithms can then operate on this smooth, averaged trajectory, focusing on the robot’s true progress rather than its wobbling head. 

Once a stable locomotion frame is established, guidance strategies can keep the robot on course. One technique is integral line-of-sight (ILOS) guidance, adapted from marine robotics \cite{ILOS} \cite{ILOS_snake_robots}. ILOS steers the robot toward a look-ahead point on the desired path while integrating the cross-track error over time. This helps counteract steady disturbances (like constant slip down a slope) and works well for long open traverses. Another technique is to use a vector field defined over the workspace \cite{guiding_vector_field} \cite{3d_vector_field}. In a vector field, every point in space has a desired heading that leads toward the path. Vector fields have no singularities and ensure global convergence, making them well-suited for navigating a cluttered crater floor with complex obstacles. 

For more complex maneuvers and mode transitions, predictive control adds anticipatory power \cite{MIT_cheetah} \cite{MPC_book} \cite{Bloch1996NonholonomicMS}. Nonlinear Model Predictive Control (NMPC) optimizes a sequence of future control inputs over a short horizon, respecting actuator limits and a model of terrain forces \cite{wang2025thrusterenhancedlocomotiondecoupledmodel} \cite{Neunert_2018} \cite{sleiman2021unifiedmpcframeworkwholebody}. With NMPC, COBRA could detect an upcoming steep slope or soft patch and adjust its gait parameters in advance. The challenge is that NMPC requires heavy computation. In practice, we mitigate this by using simplified models, warm-starting the solver, or using real-time iterative schemes. We may also employ pre-computed libraries of gaits or motions to provide good initial guesses for the optimizer. 

Learning-based methods complement these model-based approaches by offering adaptation without explicit models. For example, reinforcement learning (RL) could discover gait adjustments that handle terrains too complex to model. A practical scheme might use a hierarchical design where a high-level learned policy selects among locomotion modes or parameter ranges, while a low-level classical controller executes safe, high-rate commands \cite{10637006}. In a full deployment, we would fuse multiple components: observers that estimate terradynamic forces, feedforward gait libraries, predictive planners, and constrained learned policies. This hybrid architecture would exploit COBRA’s morphology and physics models, while adapting through learning, all within the limits of onboard computation and safety requirements \cite{salagame2024reinforcementlearningbasedmodelmatching}. 

Perception-driven behaviors can extend COBRA’s role from pure traversal to actual interaction with the environment. One example is vision-guided loco-manipulation \cite{salagame2025visionguidedlocomanipulationsnakerobot} \cite{10637029} \cite{9562974}. In this mode, COBRA uses its cameras to detect an object and estimate its pose, it then modifies its gait to approach and coil around the target as if its body were a manipulator. This lets the snake pick up or manipulate objects without a separate arm. Such a capability is valuable for tasks like autonomous sample collection or sensor placement in remote areas. Another tool is semantic terrain assessment. A vision system could classify surface types (for example, sand, gravel, bedrock, or fine dust). The planner could then build a traversability map where each terrain type has a different cost, rather than a simple occupancy map. By combining these semantic labels with the geometric map, COBRA would avoid areas that look solid but are actually soft and treacherous, and prefer routes over firmer ground. To make this work on the Moon, techniques like transfer learning or domain adaptation will be needed so that models trained on Earth images still recognize materials under lunar lighting and regolith appearance. 

On the human-robot interface side, vision-language models (VLMs) and large language models offer higher-level command interfaces. Rather than sending low-level waypoints, a scientist could give a goal like “inspect the icy boulder near the rim.” An onboard vision-language system could interpret this instruction in the context of camera imagery and prioritize tasks accordingly. In high-latency, low-bandwidth missions, such high-level grounded autonomy could greatly enhance scientific productivity. 

Finally, sensing can be improved by using COBRA’s mechanics. For example, active head stabilization uses the robot’s neck joints to counteract its own body undulation, keeping the camera more level. This reduces motion blur and rolling-shutter effects before images even reach the SLAM pipeline. Combined with the 5 locomotion frame approach, this mechanical decoupling greatly simplifies perception. It exemplifies the morpho-functional principle: the robot uses its redundant body degrees of freedom not only for movement but also to enhance its sensing performance. 

Bringing all these elements together forms a coherent roadmap for a snake-like lunar explorer. Such a robot would combine a morphology designed for low ground pressure and directional traction; materials and skins that provide passive mechanical intelligence; a visual-inertial SLAM system tuned for oscillatory motion; and control algorithms that blend predictive planning with learning. COBRA sits at the intersection of these ideas. It is more than just an alternative mobility concept; it is a synthesis of embodied intelligence, predictive autonomy, and perception-driven interaction aimed at reaching the Moon’s most scientifically valuable but mechanically challenging regions. 

Realizing this vision will require careful co-design across materials, mechanics, perception, and control. However, the potential payoff is enormous. With these capabilities, robotic explorers could access permanently shadowed craters and other extreme terrains for the first time, greatly expanding where they can go and what scientific discoveries they can achieve.

\chapter{Methodology}
\label{chap:cobra-platform}

\section{COBRA Hardware}

COBRA measures 1.7 m in length with a total mass of approximately 7 kg. The robot consists of ten identical body modules, along with dedicated head and tail units at the terminal ends. As shown in Fig.~\ref{fig:cobra_hardware} below, each module is equipped with a Dynamixel XH540-W270-R actuator capable of delivering up to 9.9 Nm of torque. Power distribution is decentralized across the platform—each module contains its own Tattu 850 mAh 4S LiPo battery regulated through LM2596 circuitry. This distributed power architecture enables approximately two hours of continuous operation without requiring centralized power management.

The body frame utilizes Markforged 3D-printed Onyx, a carbon fiber reinforced nylon material that provides an effective balance between structural integrity and mass constraints. Onboard computation is handled by an NVIDIA Jetson Orin NX processor, while perception is managed through an Intel RealSense D435i depth camera. The camera provides stereo imaging and IMU data that support state estimation and navigation tasks. For operation in dusty environments, the entire assembly can be enclosed within a continuous neoprene sheath. This protective layer prevents dust intrusion into the joints and shields internal components from abrasive regolith exposure.

\begin{figure}[t!]
    \centering
    \includegraphics[width=1\linewidth]{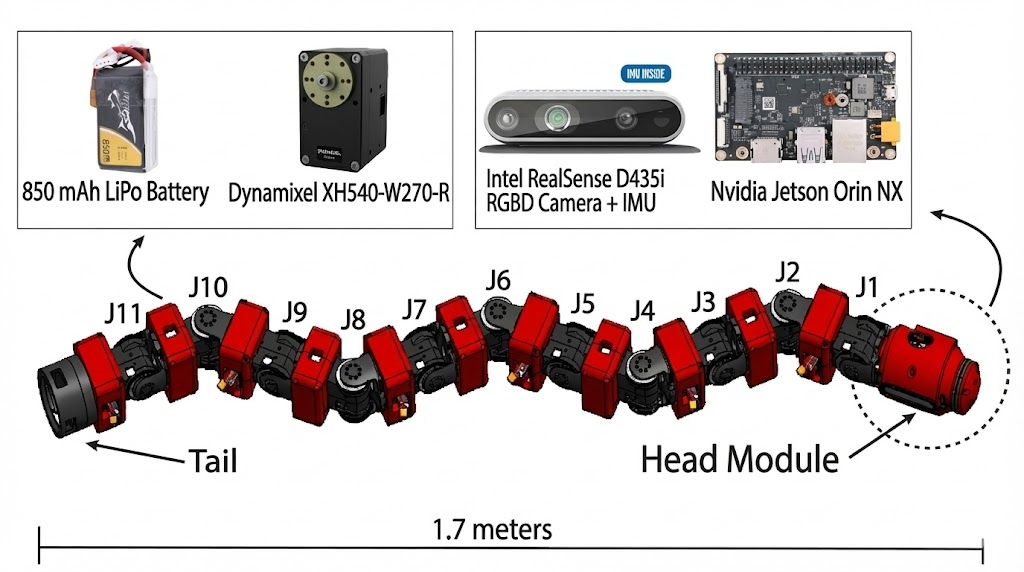}
    \caption{Illustrates COBRA robot hardware layout showing actuators, battery, depth camera, NVIDIA Jetson Orin NX.}
    \label{fig:cobra_hardware}
\end{figure}

\subsection{Head-Tail Latching Mechanism}
The head-tail latching system represents a critical structural feature that enables COBRA's multimodal locomotion capabilities. This mechanism allows the robot to actively transition from an elongated, flexible configuration into a rigid closed-loop structure suitable for tumbling gaits. The latch consists of four fin-shaped components actuated by a Dynamixel XC330 motor, which mechanically interlock the head and tail modules. The design employs purely mechanical engagement rather than magnetic coupling, thereby maintaining operational reliability in particulate-rich environments where ferromagnetic dust accumulation would compromise magnetic interface performance. When engaged, the closed-chain configuration forms a hexagonal ring that facilitates passive rolling behavior on steep terrain.

\subsection{Locomotion Plasticity}
COBRA implements six primary gait families, providing what can be characterized as "locomotion plasticity", the capacity to adapt movement strategies in response to varying terrain conditions \cite{10801515}. This adaptive capability forms the foundation of the platform's operational versatility across challenging environments.

\begin{figure}[t!]
    \centering
    \includegraphics[width=1\linewidth]{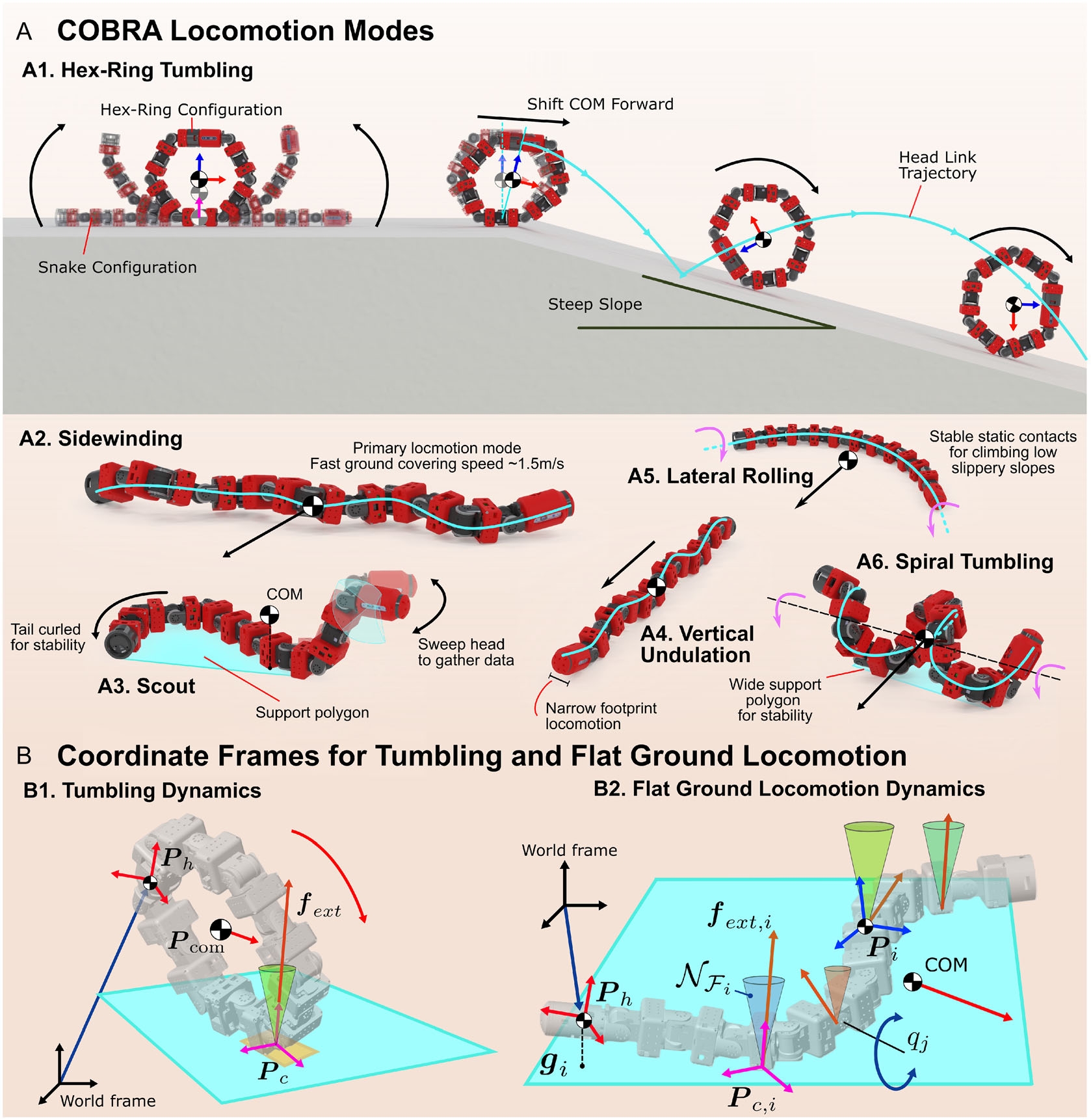}
    \caption{Overview of COBRA's locomotion plasticity. (A) Six locomotion modes: hex-ring tumbling (A1), sidewinding (A2), scout mode (A3), vertical undulation (A4), lateral rolling (A5), and spiral tumbling (A6). (B) Dynamic modeling schematic with inertial (black), head (red), link (blue), and contact (magenta) reference frames. Figure from \cite{salagame2025reduced}.}
    \label{fig:cobra_locomotion_modes}
\end{figure}

\begin{itemize}
    \item \textbf{Sidewinding:} Sidewinding serves as the default mobility mode for flat surfaces and moderate inclines \cite{sidewinding_gait}. The gait synchronizes horizontal and vertical waves along the body to produce diagonal displacement with low slip. Forward velocities reach approximately 1.5 m/s, making this the primary mode for travel between observation points.
\item \textbf{ Hex-ring tumbling:} Hex-ring tumbling activates when the robot closes into a rigid loop via the latching mechanism. The robot then functions as a passive wheel with six discrete support faces. This mode enables energy-efficient descent on slopes exceeding 30 degrees, with downhill velocities reaching approximately 5 m/s while maintaining stability through sequential face-to-face transitions.
\item \textbf{ Spiral tumbling:} Spiral tumbling forms an alternative closed-chain configuration. The robot assumes a helical shape rather than a rigid hexagon, distributing contact forces more evenly on rough, uneven ground. This provides improved control authority compared to hex-ring tumbling while still allowing gravity-assisted descent.
\item \textbf{ Lateral rolling:} Lateral rolling employs C-, S-, or J-shaped body curves to enable twisting and rolling maneuvers \cite{lateral_rolling}. This mode is effective for controlled lateral translation on moderate slopes or smooth surfaces where sidewinding proves inefficient.
\item \textbf{ Vertical undulation:} Vertical undulation compresses the robot's cross-section to approximately 10 cm by tightening vertical wave amplitudes. This allows traversal through narrow openings, fractures, or confined rock geometries—expanding the operational space to structures inaccessible to conventional wheeled vehicles.
\item \textbf{ Scout mode:} Scout mode elevates the head module while stabilizing the lower body. This aligns the sensors for terrain scanning, mapping, and situational assessment—supporting target detection, slope evaluation, and path planning before locomotion mode transitions.

\end{itemize}

\section{Software Architecture}

The control software is structured across three distinct layers to maintain separation of concerns: a ROS2 interface layer responsible for communication and state estimation, a gait library containing locomotion primitives, and a real-time CPG execution layer that generates joint commands at high frequency.

\subsection{ROS2 Integration}

The \texttt{CobraROS} class encapsulates low-level motor communication within a ROS2 node structure. During each control cycle, the node queries all active Dynamixel servos to retrieve current joint positions, velocities, and motor currents. This proprioceptive data is then integrated with base pose information obtained from an OptiTrack motion capture system to reconstruct the complete kinematic state of the robot.

Forward kinematics transforms are propagated sequentially from the tracked base frame through each of the eleven joints to determine link poses in the world coordinate frame. The system maintains pre-allocated storage for thirteen link transforms plus one center-of-mass transform to minimize memory allocation overhead during runtime. Joint positions and velocities are converted from degrees to radians to comply with ROS standard conventions before publication. The center of mass position is computed as a weighted average of the individual link positions, providing the primary input for the waypoint tracking controller.

The ROS2 layer also manages state visibility and debugging infrastructure. A TF2 transform broadcaster publishes frame transformations for all robot links, enabling real-time visualization in RViz. Joint states are published on the \texttt{joint\_states} topic, while a \texttt{bound\_state} topic provides the bounding box dimensions of the robot's current configuration. Custom \texttt{TrackingStatus} messages report the active waypoint index, tracking errors, and control outputs to support analysis and debugging. This architectural separation ensures that higher-level controllers receive consistent, well-formed pose data regardless of whether the source is motion capture, visual odometry, or simulation.

The base pose subscription implements a staleness detection mechanism that monitors the freshness of incoming pose data. If no new pose messages arrive within a configurable timeout period, the system issues warnings and eventually falls back to a default identity pose to prevent state estimation failure.

\subsection{Control Loop Timing Structure}

The system operates on a dual-rate control architecture to balance responsiveness and stability:

\begin{enumerate}
    \item \textbf{Fast Loop (CPG Execution):} The CPG layer executes at approximately 100 Hz, synchronized with the servo communication rate. At each tick, it computes and transmits new joint angle commands to all active actuators.
    \item \textbf{Slow Loop (Waypoint Tracking):} The high-level waypoint tracking controller operates at 1 Hz, making discrete navigation decisions at much longer intervals.
\end{enumerate}

This rate separation is intentional. The waypoint controller performs computationally intensive tasks such as gait mode selection, heading correction calculations, and waypoint advancement decisions. These operations require temporal integration of pose data and benefit from slower update rates that filter transient measurement noise. Running this logic at servo rate would cause the controller to react excessively to instantaneous perturbations in pose estimates, resulting in oscillatory and unstable tracking behavior.

To further attenuate measurement noise, the system applies a moving-average filter with a 10-sample window to the center-of-mass yaw angle. This filtering smooths high-frequency artifacts originating from both the motion capture system and the quaternion-to-Euler angle conversion process. The implementation includes unwrapping logic to handle the angular discontinuity at $\pm\pi$ radians, ensuring that yaw error calculations remain continuous when the robot's heading crosses the wrap-around boundary.

\section{Visual-Inertial Odometry for Real-Time State Estimation}

Autonomous waypoint tracking requires continuous, accurate estimation of the robot's pose in the world frame. While the forward kinematics pipeline described in Section [3.4.1] provides the geometric configuration of all body segments given the base pose, it requires an external reference to anchor the kinematic chain in global coordinates. For untethered operation, this reference must be obtained through onboard perception rather than external tracking systems.

\subsection{RTAB-Map Visual-Inertial SLAM Architecture}

RTAB-Map (Real-Time Appearance-Based Mapping) serves as the primary state estimation framework, providing visual-inertial odometry that tracks the robot's head pose in real-time \cite{Labb__2018}. The system fuses data from two complementary sensor modalities: stereo visual features from the Intel RealSense D435i depth camera and inertial measurements from the camera's integrated IMU. This sensor fusion approach takes advantage of the complementary characteristics of vision and inertial measurements where the visual odometry excels during steady motion with rich environmental features, while the IMU maintains tracking through rapid motions and feature-sparse intervals.

The RealSense D435i stereo camera provides stereo image pairs at 848×480 resolution and 60 Hz frame rate, with a baseline of 50 mm enabling depth estimation at 848×480 resolution and 30 Hz through triangulation. The integrated BMI055 IMU reports linear acceleration and angular velocity at 200 Hz, providing high-frequency motion measurements that bridge the relatively sparse visual updates. Both sensors are hardware-synchronized to ensure temporal alignment, which is critical for accurate sensor fusion during the dynamic sidewinding motion where head orientation can change rapidly.

RTAB-Map processes this sensor stream to generate incremental odometry estimates through a graph-based SLAM framework. The visual-inertial odometry front-end tracks feature correspondences between consecutive stereo image pairs, computes relative pose transformations, and fuses these measurements with IMU-integrated motion predictions using an Extended Kalman Filter (EKF). The back-end maintains a pose graph where nodes represent keyframe poses and edges encode relative transformation constraints derived from visual loop closures and odometry measurements. Graph optimization minimizes the cumulative error across all constraints, producing globally consistent pose estimates.

\subsection{Integration with the Control Pipeline}

The RTAB-Map odometry output provides the 6-DOF pose of the head module, consisting of position $\mathbf{p}_{\text{head}} = [x, y, z]^T \in \mathbb{R}^3$ and orientation represented as a quaternion $\mathbf{q}_{\text{head}} = [q_x, q_y, q_z, q_w]^T$. This head pose serves as the base frame input to the forward kinematics chain (Section [3.4.1]), which propagates through the joint angles to compute the complete robot configuration.

The state estimation pipeline operates as follows:

\begin{enumerate}
    \item \textbf{Visual-Inertial Odometry (30 Hz):} RTAB-Map processes stereo image pairs and IMU measurements to generate head pose estimates $\{\mathbf{p}_{\text{head}}, \mathbf{q}_{\text{head}}\}$ at the camera frame rate.
    
    \item \textbf{Joint State Acquisition (100 Hz):} Dynamixel XH540 servo actuators report absolute joint angles $\mathbf{q} \in \mathbb{R}^{11}$ through their internal encoders at 100 Hz, providing high-frequency kinematic state updates.
    
    \item \textbf{Forward Kinematics Propagation (100 Hz):} The most recent head pose is combined with current joint angles to compute world-frame positions $\mathbf{P}_{\text{links}} \in \mathbb{R}^{3 \times 13}$ for all body segments via the forward kinematics mapping $\mathcal{F}(\mathbf{X}_b)$ (Equation 3.6).
    
    \item \textbf{Center of Mass Estimation (100 Hz):} The center of mass position $\mathbf{P}_{\text{com}}$ and virtual chassis orientation $\mathbf{R}_{\text{com}}$ are derived from the link positions using the mass-weighted averaging and SVD-based orientation computation described in Section [X.X].
    
    \item \textbf{Feedback Control (1 Hz):} The CoM pose $\{\mathbf{P}_{\text{com}}, \mathbf{R}_{\text{com}}\}$ is sampled at 1 Hz to provide the feedback signal for the high-level trajectory tracking controller, which computes steering corrections $\delta$ based on waypoint errors.
\end{enumerate}

This hierarchical update structure decouples the high-frequency kinematic state propagation (100 Hz) from the lower-rate visual odometry updates (30 Hz) and control decisions (1 Hz).

\subsection{Coordinate Frame Transformations}

RTAB-Map outputs pose estimates in a camera-centric coordinate frame that must be transformed to the robot's kinematic reference frame. The RealSense D435i camera is rigidly mounted within the head module with a fixed transformation $\mathbf{T}_{\text{head}}^{\text{camera}} \in SE(3)$ determined through offline calibration. This transformation accounts for both the positional offset between the camera optical center and the head frame origin, as well as the rotational alignment between coordinate conventions.

The complete transformation chain from camera odometry to base frame pose is:

\begin{equation}
\mathbf{T}_{\text{base}}^{\text{world}} = \mathbf{T}_{\text{camera}}^{\text{world}} \cdot (\mathbf{T}_{\text{head}}^{\text{camera}})^{-1} \cdot \mathbf{T}_{\text{base}}^{\text{head}}
\end{equation}

where $\mathbf{T}_{\text{camera}}^{\text{world}}$ is the RTAB-Map odometry output, $(\mathbf{T}_{\text{head}}^{\text{camera}})^{-1}$ is the inverse of the fixed camera-to-head calibration, and $\mathbf{T}_{\text{base}}^{\text{head}}$ is the kinematic transformation from head to base frame defined in the forward kinematics (Equation X.X). The resulting base pose $\mathbf{T}_{\text{base}}^{\text{world}}$ initializes the forward kinematics chain for the current time step.

\subsection{Computational Architecture}
The visual-inertial SLAM system executes on the Nvidia Jetson Orin Nano embedded platform housed in the robot's head module. RTAB-Map is implemented as a ROS 2 (Robot Operating System) node, with odometry estimates published as \texttt{nav\_msgs/Odometry} messages. Forward kinematics and center of mass estimation nodes subscribe to both odometry (30 Hz) and joint state (100 Hz) topics, ensuring the CoM pose reflects the most current available information. Between visual updates, the CoM position evolves based on joint motion alone, which is valid for the short inter-frame intervals ($\sim 33\,\text{ms}$) given the slow CoM velocities during sidewinding (typically $< 0.5\,\text{m/s}$). The high-level trajectory tracking controller operates at 1 Hz, generating CPG parameter updates that are transmitted to the gait generation layer.

\section{Kinematic Computations}

The state estimation pipeline transforms raw hardware measurements into a complete geometric representation of the robot's configuration. This process involves forward kinematics propagation, virtual chassis frame construction, and center-of-mass calculation to derive a reduced-order representation of the multi-DOF system \cite{salagame2025reduced} \cite{salagame2025optimaltrajectoryplanningvertically}.

\subsection{Forward Kinematics Chain}

COBRA's kinematic structure consists of a tracked base frame followed by thirteen body segments connected through eleven actuated joints arranged in an alternating pitch-yaw configuration. The forward kinematics solver computes the world-frame pose of each link through successive homogeneous transformations.

The robot state relative to the base frame is defined by the vector $\mathbf{X}_b = [\mathbf{p}_b^T, \boldsymbol{\Phi}_b^T, \mathbf{q}^T]^T$, where $\mathbf{p}_b$ represents the position of the base frame, $\boldsymbol{\Phi}_b$ describes its orientation using XYZ-Euler angles, and $\mathbf{q}$ denotes the joint angles of each link. The base transformation is defined by the tracked pose from the motion capture system:



\begin{equation}
\mathbf{T}_{\text{base}} = \begin{bmatrix}
\mathbf{R}_{\text{base}} & \mathbf{p}_{\text{base}} \\
\mathbf{0}^T & 1
\end{bmatrix} \in SE(3)
\end{equation}
where $\mathbf{R}_{\text{base}} \in SO(3)$ is the rotation matrix derived from the base quaternion $\mathbf{q}_{\text{base}} = [q_x, q_y, q_z, q_w]^T$, and $\mathbf{p}_{\text{base}} = [x, y, z]^T$ is the position vector.

The transformation from base to head incorporates a translation along the local $x$-axis by the head length $H = 0.1565$ m, followed by a $+90^\circ$ rotation about the $x$-axis:
\begin{equation}
\mathbf{T}_{\text{head}} = \mathbf{T}_{\text{base}} \cdot \text{Trans}(-H, 0, 0) \cdot \mathbf{R}_x(+90^\circ)
\end{equation}

Subsequent links propagate through the kinematic chain using the general form:
\begin{equation}
\mathbf{T}_{i+1} = \mathbf{T}_i \cdot \mathbf{R}_z(-q_i) \cdot \text{Trans}(-L, 0, 0) \cdot \mathbf{R}_x(\pm 90^\circ)
\end{equation}
where $L = 0.1230$ m is the standard link length, $q_i$ is the joint angle in radians, and the sign of the $x$-axis rotation alternates between consecutive joints to create the pitch-yaw sequence. The negative sign in $\mathbf{R}_z(-q_i)$ accounts for the joint angle convention used in the hardware interface."

For computational efficiency, the rotation matrices $\mathbf{R}_x(\pm 90^\circ)$ and translation matrix \allowbreak $\text{Trans}(-L, 0, 0)$ are precomputed as static constants. The $z$-axis rotation is computed inline for each joint:

\begin{equation}
\mathbf{R}_z(-q_i) = \begin{bmatrix}
\cos(-q_i) & -\sin(-q_i) & 0 \\
\sin(-q_i) & \cos(-q_i) & 0 \\
0 & 0 & 1
\end{bmatrix}
\end{equation}

Using forward kinematics, the position of each link in the world frame, $\mathbf{p}_i$, is computed. The resulting $3 \times N$ matrix is given by:
\begin{equation}
\mathbf{P}_{\text{links}} = \mathcal{F}(\mathbf{X}_b)
\end{equation}
which encapsulates the positions of all $N = 13$ links (base + head + 11 body links) in the world frame, where $\mathcal{F}$ represents the forward kinematics mapping.

From each homogeneous transformation $\mathbf{T}_i$, both world-frame and parent-frame poses are extracted:
\begin{align}
\mathbf{p}_i^{\text{world}} &= \mathbf{T}_i[1{:}3, 4] \\
\mathbf{R}_i^{\text{world}} &= \mathbf{T}_i[1{:}3, 1{:}3] \\
\mathbf{p}_i^{\text{parent}} &= \mathbf{T}_{\text{rel},i}[1{:}3, 4] \\
\mathbf{R}_i^{\text{parent}} &= \mathbf{T}_{\text{rel},i}[1{:}3, 1{:}3]
\end{align}
where $\mathbf{T}_{\text{rel},i}$ represents the relative transformation between consecutive frames.

\subsection{Virtual Chassis Frame Construction}

The virtual chassis (VC) frame provides a body-fixed coordinate system that tracks the robot's overall orientation while remaining horizontal relative to the world frame \cite{salagame2025reduced}. This floating reference frame serves as the reference for computing body-relative quantities such as bounding boxes and contact states, and remains largely stable and independent of individual link oscillations during gait execution \cite{Rollinson01122012} \cite{Rollinson2011StateEF}.

The VC frame is constructed by first computing the average forward direction across all link frames:
\begin{equation}
\bar{\mathbf{x}} = \frac{1}{N} \sum_{i=0}^{12} \mathbf{R}_i^{\text{world}} \begin{bmatrix} 1 \\ 0 \\ 0 \end{bmatrix}
\end{equation}
where $N = 13$ is the total number of frames (base + head + 11 links).

This average direction is then projected onto the horizontal plane and normalized to obtain the VC $x$-axis:
\begin{equation}
\hat{\mathbf{x}}_{\text{com}} = \frac{\bar{\mathbf{x}} - (\bar{\mathbf{x}} \cdot \hat{\mathbf{z}}) \hat{\mathbf{z}}}{\|\bar{\mathbf{x}} - (\bar{\mathbf{x}} \cdot \hat{\mathbf{z}}) \hat{\mathbf{z}}\|}
\end{equation}
where $\hat{\mathbf{z}} = [0, 0, 1]^T$ is the world vertical axis.

The VC $y$-axis is obtained through the cross product to complete a right-handed orthonormal basis:
\begin{equation}
\hat{\mathbf{y}}_{\text{com}} = \hat{\mathbf{z}} \times \hat{\mathbf{x}}_{\text{com}}
\end{equation}

The virtual chassis rotation matrix is then assembled as:
\begin{equation}
\mathbf{R}_{\text{com}} = \begin{bmatrix} \hat{\mathbf{x}}_{\text{com}} & \hat{\mathbf{y}}_{\text{com}} & \hat{\mathbf{z}} \end{bmatrix} \in SO(3)
\end{equation}

This frame construction ensures that the VC yaw angle accurately represents the robot's heading direction while filtering out pitch and roll variations from individual link orientations.

\subsubsection{Center of Mass Calculation}

The center of mass position is computed as the mass-weighted average of all link positions:
\begin{equation}
\mathbf{P}_{\text{com}} = \frac{\sum_{i=0}^{12} m_i \mathbf{p}_i^{\text{world}}}{\sum_{i=0}^{12} m_i}
\end{equation}
where the individual link masses are:
\begin{equation}
\mathbf{m} = [m_{\text{base}}, m_{\text{head}}, m_1, \ldots, m_{10}, m_{\text{tail}}] = [0.25, 0.25, 0.50, \ldots, 0.50] \text{ kg}
\end{equation}

The reciprocal total mass is precomputed to replace division with multiplication during the summation. The center of mass position $\mathbf{P}_{\text{com}}$ serves as the primary control point for the waypoint tracking controller, as it represents the robot's effective position for navigation purposes.

\begin{figure}[t!]
    \centering
    \includegraphics[width=1\linewidth]{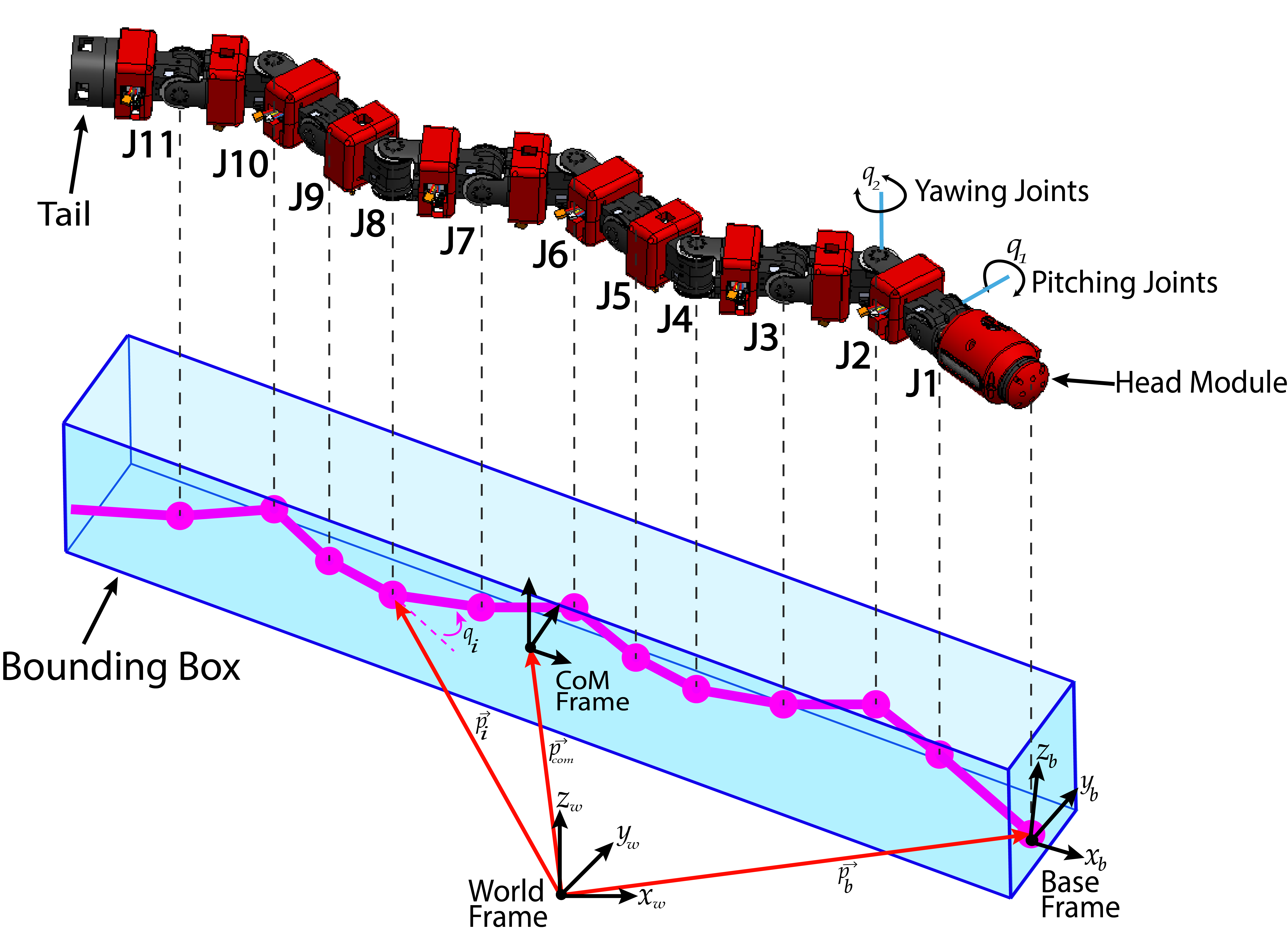}
    \caption{This image shows the reduced-order COBRA model with the bounding box, center-of-mass (CoM) frame, and base frame expressed in the world coordinate frame. The model uses a consistent naming convention where Jx denotes joints (x = 1-11) and Ly denotes links (y = 1-10), with coordinate frames defined for yawing and pitching joints.}
    \label{fig:3d_cobra}
\end{figure}

\subsection{Bounding Box Computation}

The bounding box dimensions are computed in the virtual chassis frame to provide a body-relative measure of the robot's spatial extent. To further simplify the reduced-order model, this bounding box abstracts the robot's internal configuration, focusing instead on its external spatial extent for path planning in constrained environments. Each link position is first transformed into the VC frame:
\begin{equation}
\tilde{\mathbf{P}}_{\text{links}} = \mathbf{P}_{\text{links}} - \begin{bmatrix} \mathbf{P}_{\text{com}}^T, \ldots, \mathbf{P}_{\text{com}}^T \end{bmatrix}
\end{equation}
\begin{equation}
\bar{\mathbf{P}}_{\text{links}} = \mathbf{R}_{\text{com}}^T \tilde{\mathbf{P}}_{\text{links}}
\end{equation}

The axis-aligned bounding box is then determined by finding the minimum and maximum coordinates along each axis. The extents along each axis are computed as:
\begin{equation}
\Delta_k = \frac{1}{2} \left[ \max(\bar{\mathbf{P}}_{\text{links}}\hat{k}) - \min(\bar{\mathbf{P}}_{\text{links}}\hat{k}) \right]
\end{equation}
where $k \in \{x, y, z\}$. This yields the half-dimensions of the bounding box along the $\hat{\mathbf{x}}$, $\hat{\mathbf{y}}$, and $\hat{\mathbf{z}}$ axes.

The bounding box span vector includes 100 mm of padding along each dimension:
\begin{equation}
\mathbf{b}_{\text{span}} = \begin{bmatrix}
2\Delta_x + 0.1 \\
2\Delta_y + 0.1 \\
2\Delta_z + 0.1
\end{bmatrix} \text{ m}
\end{equation}

This bounding box representation provides a compact description of the robot's current configuration and can be used for collision checking or workspace analysis during trajectory planning.

\section{Central Pattern Generator Architecture}

The low-level motion generation for COBRA employs a network of coupled nonlinear oscillators that produces smooth, coordinated joint trajectories without requiring explicit trajectory planning for each degree of freedom. This approach is inspired by the biological central pattern generators found in vertebrate spinal cords, which are capable of generating rhythmic locomotion patterns independent of descending brain commands \cite{salagame2025reduced}. The CPG framework enables the generation of complex gait primitives through the manipulation of a small set of high-level parameters, while the coupled oscillator dynamics ensure smooth inter-joint coordination \cite{9992457}.

\subsection{Coupled Oscillator Dynamics}

Each of COBRA's eleven actuated joints is governed by an individual Hopf oscillator characterized by two state variables: the phase angle $\theta_i$, which defines the instantaneous position within the oscillation cycle, and the amplitude $r_i$, which determines the magnitude of the oscillation. The complete internal state of the CPG system is represented by the vector $\mathbf{X}_{\text{cpg}} = [\boldsymbol{\theta}^T, \mathbf{r}^T, \dot{\mathbf{r}}^T]^T$, where $\boldsymbol{\theta} \in \mathbb{R}^{11}$, $\mathbf{r} \in \mathbb{R}^{11}$, and $\dot{\mathbf{r}} \in \mathbb{R}^{11}$ represent the phase, amplitude, and rate of change of amplitude for all oscillators, respectively.

The oscillators are not independent but rather coupled to their immediate neighbors through phase dynamics, enabling the propagation of traveling waves along the robot's kinematic chain. The phase evolution is governed by:

\begin{equation} \label{eq:phase_dynamics}
\dot{\boldsymbol{\theta}} = \mu \mathbf{A} \boldsymbol{\theta} - \mu \mathbf{B} \boldsymbol{\Delta\phi} + \boldsymbol{\omega}
\end{equation}

where $\mathbf{A} \in \mathbb{R}^{11 \times 11}$ is a tridiagonal coupling matrix that enforces nearest-neighbor interactions, $\mathbf{B} \in \mathbb{R}^{11 \times 10}$ maps the desired relative phase offsets to the phase dynamics, $\boldsymbol{\Delta\phi} \in \mathbb{R}^{10}$ contains the phase differences between consecutive joints, $\boldsymbol{\omega} \in \mathbb{R}^{11}$ specifies the natural frequencies, and $\mu$ is a coupling strength parameter that regulates the intensity of phase synchronization.

The coupling matrices are defined as:

\begin{equation}
\mathbf{A} = 
\begin{bmatrix}
1 & 1 & & & \\
1 & -2 & \ddots & & \\
& \ddots & \ddots & \ddots & \\
& & \ddots & -2 & 1 \\
& & & 1 & 1
\end{bmatrix}, \quad
\mathbf{B} = 
\begin{bmatrix}
1 & & & \\
-1 & \ddots & & \\
& \ddots & \ddots & \\
& & -1 & 1
\end{bmatrix}
\end{equation}

The tridiagonal structure of $\mathbf{A}$ ensures that each oscillator is coupled exclusively to its immediate neighbors in the kinematic chain, creating a mechanism for local synchronization that produces globally coordinated behavior. The relative phase vector $\boldsymbol{\Delta\phi}$ encodes the desired phase relationships between adjacent joints, which directly determines the spatial wavelength of the resulting body wave.

The amplitude dynamics are formulated as a critically damped second-order system:

\begin{equation} \label{eq:amplitude_dynamics}
\ddot{\mathbf{r}} = -\gamma^2 \mathbf{r} + \gamma^2 \mathbf{a}_{\text{des}} - \gamma \dot{\mathbf{r}}
\end{equation}

where $\mathbf{a}_{\text{des}} \in \mathbb{R}^{11}$ represents the desired amplitude vector and $\gamma$ is the natural frequency of the amplitude dynamics. The critical damping condition ensures smooth, monotonic convergence to target amplitudes without overshoot, which is particularly important when the high-level controller modulates gait parameters during execution.

The complete state update model can be expressed in compact matrix form:

\begin{equation}
\begin{bmatrix}
\dot{\boldsymbol{\theta}} \\
\dot{\mathbf{r}} \\
\ddot{\mathbf{r}}
\end{bmatrix} = 
\begin{bmatrix}
\mu \mathbf{A} & \mathbf{0} & \mathbf{0} \\
\mathbf{0} & \mathbf{0} & \mathbf{I} \\
\mathbf{0} & -\gamma^2 \mathbf{I} & -\gamma \mathbf{I}
\end{bmatrix}
\begin{bmatrix}
\boldsymbol{\theta} \\
\mathbf{r} \\
\dot{\mathbf{r}}
\end{bmatrix} +
\begin{bmatrix}
\mathbf{0} & \mathbf{I} & -\mu \mathbf{B} \\
\mathbf{0} & \mathbf{0} & \mathbf{0} \\
\gamma^2 \mathbf{I} & \mathbf{0} & \mathbf{0}
\end{bmatrix}
\begin{bmatrix}
\mathbf{a}_{\text{des}} \\
\boldsymbol{\omega} \\
\boldsymbol{\Delta\phi}
\end{bmatrix}
\end{equation}

The desired joint angles generated by the CPG are obtained through:

\begin{equation}
\mathbf{q}_{\text{des}} = \mathbf{r} \odot \sin(\boldsymbol{\theta}) + \mathbf{b}
\end{equation}

where $\odot$ denotes element-wise multiplication and $\mathbf{b} \in \mathbb{R}^{11}$ is a bias vector that allows joints to oscillate around non-zero equilibrium positions.

\subsection{Gait Parameterization}

COBRA features an alternating pitch-yaw joint configuration in which odd-indexed joints ($J_1, J_3, J_5, J_7, J_9, J_{11}$) control sagittal plane motion (pitching), while even-indexed joints ($J_2, J_4, J_6, J_8, J_{10}$) control transverse plane motion (yawing). This kinematic structure enables the generation of three-dimensional body undulations through appropriate parameter selection.

A complete gait is specified by four parameter vectors:
\begin{itemize}
    \item \textbf{Frequency ($\boldsymbol{\omega} \in \mathbb{R}^{11}$):} Specifies temporal oscillation frequency. For synchronized gaits such as sidewinding \cite{bhattachan2025cobra}, all elements are set to a common value $\omega_0$, ensuring phase-locked coordination.

\item \textbf{Phase ($\boldsymbol{\phi} \in \mathbb{R}^{11}$):} Defines absolute joint phases, with relative differences $\boldsymbol{\Delta\phi}$ determining spatial gait structure. Sidewinding gaits employ orthogonal traveling waves in sagittal and transverse planes through phase relationships maintaining a $90^\circ$ offset between vertical and horizontal components.

\item \textbf{Amplitude ($\mathbf{a}_{\text{des}} \in \mathbb{R}^{11}$):} Controls oscillation magnitude, directly influencing the robot's footprint and locomotion efficiency. In sidewinding, pitching joints typically employ smaller amplitudes ($\approx 15^\circ$) compared to yawing joints ($\approx 45^\circ$), creating the characteristic lifting pattern wherein body portions are elevated while others maintain ground contact.

\item \textbf{Bias ($\mathbf{b} \in \mathbb{R}^{11}$):} Establishes equilibrium positions for joint oscillations. Non-zero biases enable asymmetric gaits and postural adjustments for non-neutral locomotion configurations.
\end{itemize}

This parameterization provides a compact interface for gait design, where complex multi-joint coordination emerges from relatively few high-level parameters. The coupled oscillator dynamics ensure smooth transitions between parameter sets, enabling online gait modulation for adaptive control \cite{salagame2025reduced, 10637006}.





\section{Steering Through Amplitude Modulation}

The waypoint tracking controller generates steering commands by modulating the amplitude distribution of horizontal-axis joints (yawing joints: $J_2, J_4, J_6, J_8, J_{10}$) while maintaining constant amplitudes for vertical-axis joints (pitching joints: $J_1, J_3, J_5, J_7, J_9, J_{11}$) \cite{bhattachan2025cobra}. This approach exploits the fundamental asymmetry principle: differential lateral displacement between anterior and posterior body segments generates a net angular velocity, enabling curved trajectory execution within the sidewinding locomotion framework.

\subsection{Asymmetric Amplitude Distribution}

When the controller computes a steering correction $\delta$ based on heading error, it applies asymmetric modifications to the baseline horizontal amplitudes. The corrected amplitude for each horizontal joint is computed as:

\begin{equation}
a_i^{\text{corrected}} = a_i^{\text{nominal}} + \alpha_i \cdot \delta + \delta_{\text{offset}}
\end{equation}

where $a_i^{\text{nominal}}$ is the baseline sidewinding amplitude for joint $i$, $\alpha_i$ is a joint-specific weighting coefficient, $\delta$ is the steering correction magnitude, and $\delta_{\text{offset}}$ is a bias term that compensates for systematic drift in the nominal gait.

The weighting coefficients are structured to create a gradient along the body: anterior joints receive amplification proportional to their distance from the center of mass, while posterior joints receive corresponding attenuation. Specifically, the weighting distribution follows:

\begin{equation}
\boldsymbol{\alpha} = [\alpha_2, \alpha_4, \alpha_6, \alpha_8, \alpha_{10}]^T = [2.0, 1.0, 0.0, -1.0, -2.0]^T
\end{equation}

This gradient structure ensures that Joint 2 (anterior-most horizontal joint) receives twice the correction magnitude of Joint 4, while maintaining zero correction at the central joint ($J_6$), and applying opposing corrections of increasing magnitude toward the posterior joints. The linear gradient is crucial for generating smooth, continuous curvature rather than abrupt directional changes that could destabilize the contact pattern \cite{jiang2024hierarchical}.

\subsection{Mechanism of Lateral Thrust Asymmetry}

The asymmetric amplitude distribution creates unbalanced lateral thrust through differential arc lengths traced by body segments during each gait cycle. When anterior joints execute larger horizontal oscillations than posterior joints, the front portion of the robot sweeps through a wider lateral range, generating greater displacement perpendicular to the nominal forward direction. This imbalance produces a moment about the vertical axis passing through the robot's center of mass, causing the heading angle to rotate toward the side with reduced amplitude.

The instantaneous turning rate can be approximated by considering the velocity differential between the anterior and posterior contact regions. For small steering corrections where $|\delta| \ll a^{\text{nominal}}$, the angular velocity is approximately:

\begin{equation}
\dot{\theta} \approx \frac{k_{\text{turn}} \cdot \delta}{L_{\text{postural}}}
\end{equation}

where $k_{\text{turn}}$ is an empirically determined gain coefficient that depends on gait frequency and nominal amplitude, and $L_{\text{postural}}$ is the characteristic postural width of the sidewinding gait (the lateral extent of the body wave).

\subsection{Safety Constraints and Saturation}

To prevent joint commands from exceeding hardware limits or entering kinematic singularities, amplitude corrections are subject to strict saturation constraints. The corrected amplitude for each joint must satisfy:

\begin{equation}
a_i^{\text{corrected}} = \text{clip}(a_i^{\text{nominal}} + \alpha_i \cdot \delta + \delta_{\text{offset}}, \, a_{\min}, \, a_{\max})
\end{equation}

where $a_{\min}$ and $a_{\max}$ represent the minimum and maximum permissible oscillation amplitudes, typically set to $5^\circ$ and $50^\circ$ respectively for COBRA's Dynamixel XH540 actuators. These bounds ensure that the corrected amplitudes remain within the mechanical range of motion and maintain sufficient clearance from kinematic limits.

Additionally, the steering correction magnitude $\delta$ itself is bounded to prevent excessive control actions:

\begin{equation}
\delta = \text{clip}(\delta_{\text{raw}}, \, -\delta_{\max}, \, +\delta_{\max})
\end{equation}

where $\delta_{\max}$ is the maximum allowable correction, typically set to $15^\circ$ to balance responsiveness with stability.

\subsection{Bias Correction for Systematic Drift}

The offset term $\delta_{\text{offset}}$ addresses systematic heading drift that can arise from asymmetries in the physical robot (e.g., uneven joint friction, slight mass distribution imbalances, or calibration errors). This bias is determined through an initial calibration procedure where the robot executes nominally straight sidewinding trajectories, and the persistent heading drift rate is measured. The offset is then computed as:

\begin{equation}
\delta_{\text{offset}} = -K_{\text{bias}} \cdot \langle \dot{\theta}_{\text{drift}} \rangle
\end{equation}

where $\langle \dot{\theta}_{\text{drift}} \rangle$ is the time-averaged drift rate measured during straight-line motion, and $K_{\text{bias}}$ is a calibration gain. This feedforward compensation ensures that when $\delta = 0$ (no intentional steering), the robot maintains a straight trajectory rather than gradually curving due to hardware asymmetries.

The complete amplitude modulation framework integrates with the CPG architecture described in Section [X.X], where the corrected amplitudes $\mathbf{a}^{\text{corrected}}$ replace the nominal amplitude vector $\mathbf{a}_{\text{des}}$ in the amplitude dynamics (Equation \ref{eq:amplitude_dynamics}). The critically damped second-order amplitude response ensures that changes in the correction signal $\delta$ produce smooth, continuous variations in the actual joint amplitudes, preventing sudden transients that could disrupt the sidewinding contact pattern.

\section{State Machine Structure}

The tracking controller operates as a three-mode state machine:

\begin{itemize}
    \item \textbf{Sidewinding Mode:} The default state. The robot moves toward the waypoint, making continuous steering adjustments.
    \item \textbf{Turning Mode:} If the yaw error gets too large (exceeding 40 degrees), the controller drops sidewinding and switches to a dedicated ``turn-in-place'' gait. It stays here until the heading error falls below a tighter threshold (30 degrees), then switches back.
    \item \textbf{Approach Mode:} Behavior changes when the robot gets within 35 centimeters of the waypoint. The logic shifts from using relative error (direction \textit{to} the waypoint) to absolute yaw error (alignment \textit{with} the waypoint's desired heading). It will even perform in-place corrections here to nail the final orientation.
\end{itemize}

To make these transitions seamless, we use a blending function for the combined yaw error. Far from the target, the robot cares about pointing \textit{towards} it (relative yaw error); close up, it cares about matching the \textit{final heading} (absolute yaw error). We use smoothstep transitions to interpolate between these two goals based on distance, preventing abrupt signal changes that could trigger oscillations.

This is implemented as a trapezoidal weight function where w = 1.0 at the waypoint (prioritizing absolute yaw error) and w = 0.0 beyond 1.5 meters (prioritizing relative yaw error). The combined yaw error is computed as:
\[
\text{combined\_yaw\_error} = w \cdot \text{yaw\_abs} + (1 - w) \cdot \text{yaw\_rel}
\]

where the weight w(d) varies with distance d according to:

\[
w(d) = \begin{cases}
1.0 & \text{if } d = 0 \\
\\
1.0 - 0.5 \cdot s\!\left(\frac{d - 0}{0.5 - 0}\right) & \text{if } 0 < d \leq 0.5 \\
\\
0.5 & \text{if } 0.5 < d \leq 1.0 \\
\\
0.5 - 0.5 \cdot s\!\left(\frac{d - 1.0}{1.5 - 1.0}\right) & \text{if } 1.0 < d \leq 1.5 \\
\\
0.0 & \text{if } d > 1.5
\end{cases}
\]

The smoothstep function s(t) ensures C¹ continuity at transition boundaries:
\[
s(t) = 3t^2 - 2t^3, \quad t \in [0, 1]
\]

\begin{figure}[htb] 
    \centering
    \includegraphics[width=1\linewidth]{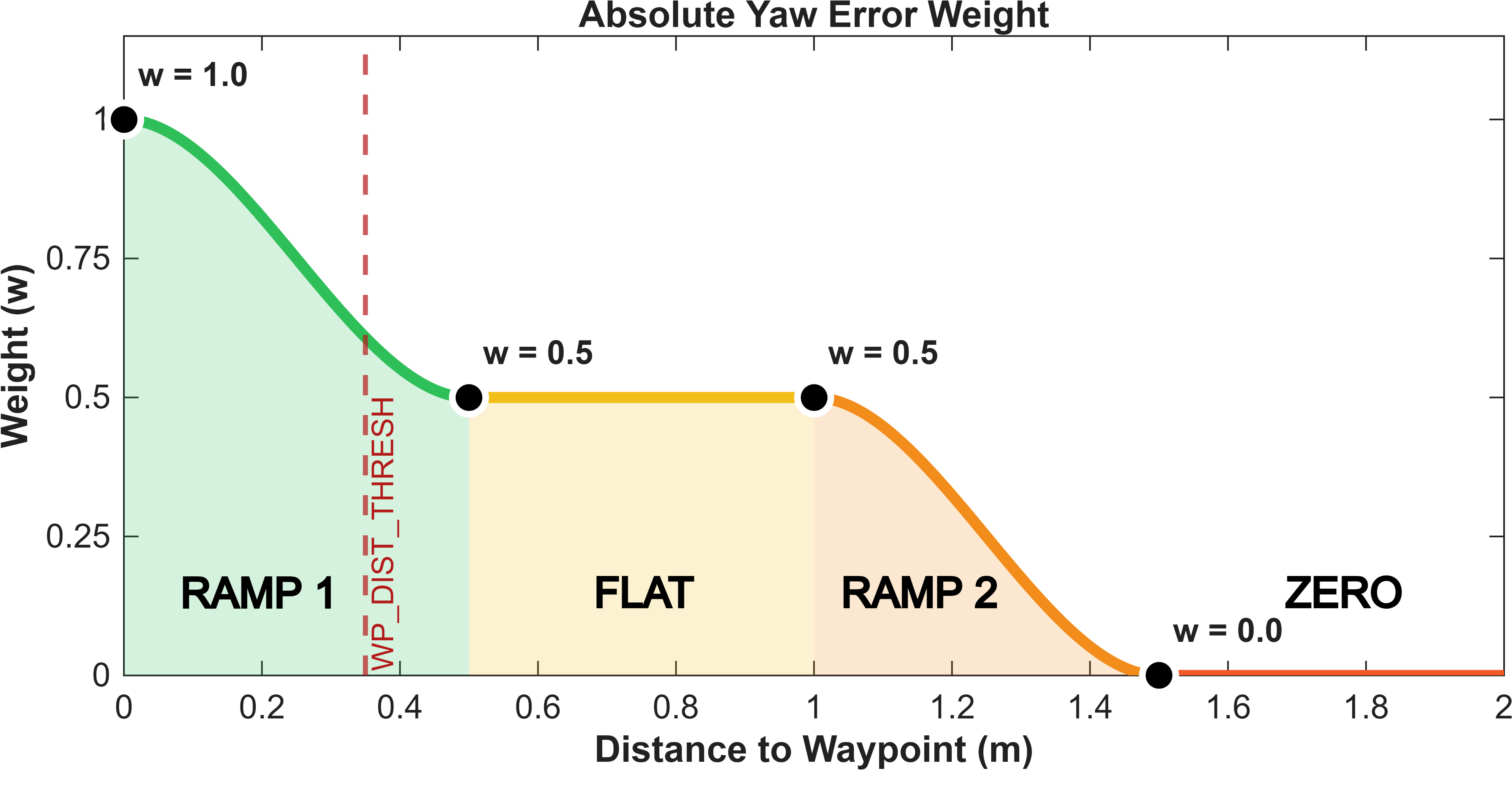}
    \caption{Illustrates the absolute yaw error weight's contribution due to the net yaw error based on distance from the next waypoint.}
    \label{fig:yaw_error_weight_plot_1}
\end{figure}

\begin{figure}[htb]
    \centering
    \includegraphics[width=1\linewidth]{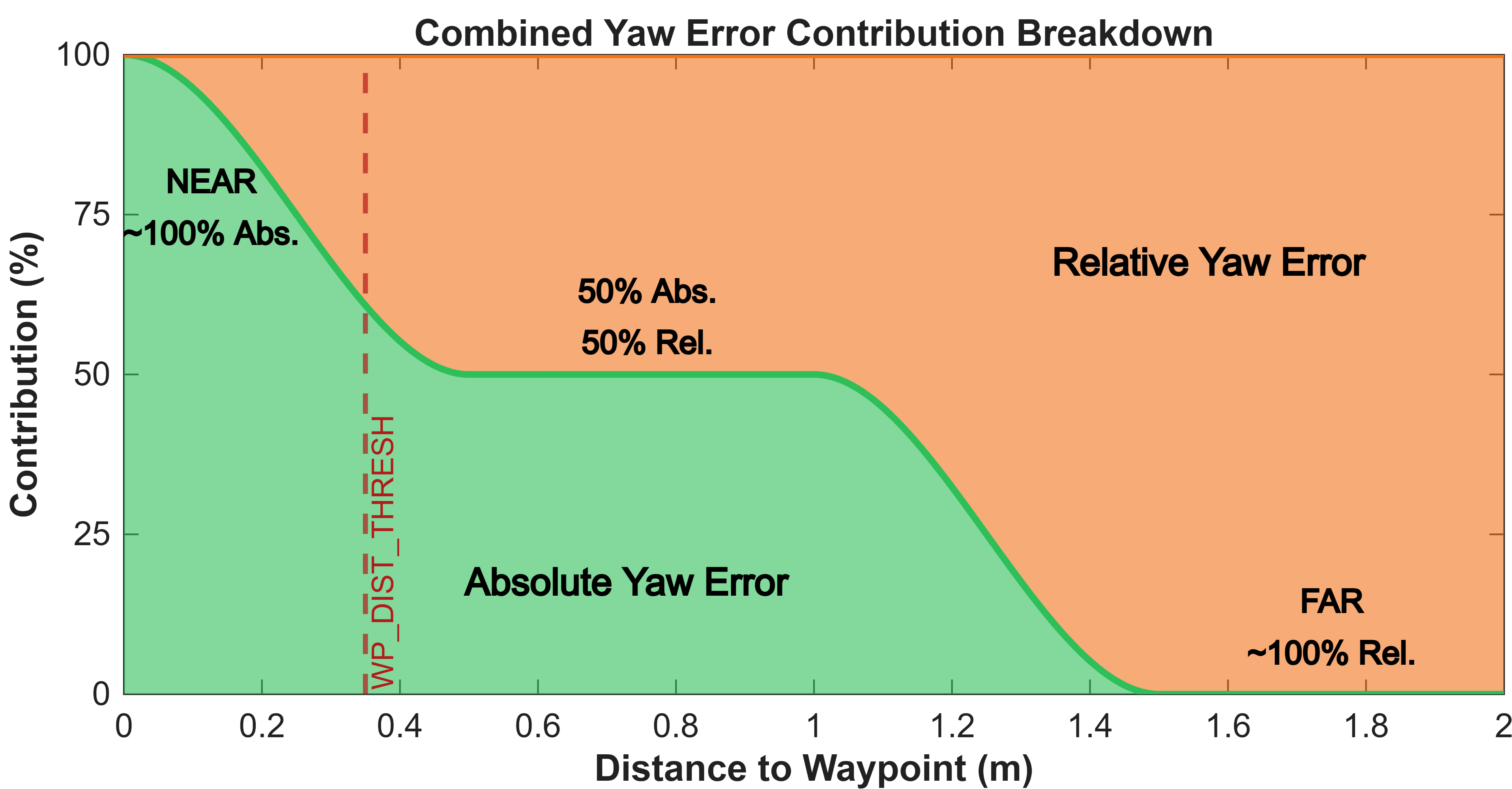}
    \caption{Illustrates the combined yaw error contribution due to absolute and relative yaw error each based on distance from the next waypoint}
    \label{fig:yaw_error_weight_plot_2}
\end{figure}

\clearpage  

\begin{center}
\textbf{Algorithm 1:} Closed-Loop Waypoint Tracking with CPG-Based Sidewinding Locomotion
\end{center}

\begin{algorithmic}[1]
\small

\Require Waypoints $\mathcal{W} = \{(\mathbf{p}_i, \psi_i)\}_{i=1}^{N}$, thresholds $d_{\text{th}}$, $\psi_{\text{th}}^{\text{wp}}$, $\psi_{\text{th}}^{\text{turn}}$, $\delta_0$, $\delta_{\text{lim}}$

\State \textbf{Initialize:} $k \gets 1$, $\text{mode} \gets 1$
\State \Call{StartCPG}{$\mathbf{a}_{\text{sw}}, \boldsymbol{\omega}, \boldsymbol{\phi}_{\text{sw}}$}

\While{$k \leq N$}
    \State $(\mathbf{p}, \psi) \gets$ \Call{GetCOMPose}{ }
    \Comment{Compute errors}
    \State $d_e \gets \|\mathbf{p}_k - \mathbf{p}\|_2$
    \State $\psi_e^{\text{rel}} \gets \text{wrap}\bigl(\text{atan2}(p_{k,y} - p_y, p_{k,x} - p_x) - \psi\bigr)$
    \State $\psi_e^{\text{abs}} \gets \text{wrap}(\psi_k - \psi)$
    \State $\psi_e \gets w(d_e) \cdot \psi_e^{\text{abs}} + (1 - w(d_e)) \cdot \psi_e^{\text{rel}}$
    
\If{$d_e > d_{\text{th}}$} \Comment{FAR branch}
    \State \Call{ProcessFarBranch}{$\psi_e$}
\Else \Comment{NEAR branch}
    \State \Call{ProcessNearBranch}{$\psi_e, k$}
\EndIf
\EndWhile
\State \Call{StopCPG}{ }

\Statex
\noindent\rule{\linewidth}{0.4pt}
\Statex

\Procedure{ProcessFarBranch}{$\psi_e$}
    \If{$\text{mode} = 1$}
        \If{$|\psi_e| \leq \psi_{\text{th}}^{\text{turn}}$}
            \State $\delta \gets \delta_0 + \text{clamp}(K_p \psi_e, -\delta_{\text{lim}}, +\delta_{\text{lim}})$
            \State $\mathbf{a} \gets$ \Call{ModifyAmplitudes}{$\mathbf{a}_{\text{sw}}, \delta$}
            \State \Call{UpdateCPG}{$\mathbf{a}, \boldsymbol{\omega}, \boldsymbol{\phi}_{\text{sw}}$}
        \ElsIf{$\psi_e > +\psi_{\text{th}}^{\text{turn}}$ \textbf{or} $\delta > +20^{\circ}$}
            \State $\text{mode} \gets 2$
            \State \Call{UpdateCPG}{$\mathbf{a}_{\text{turn}}, \boldsymbol{\omega}, \boldsymbol{\phi}_{\text{L}}$}
        \ElsIf{$\psi_e < -\psi_{\text{th}}^{\text{turn}}$ \textbf{or} $\delta < -20^{\circ}$}
            \State $\text{mode} \gets 2$
            \State \Call{UpdateCPG}{$\mathbf{a}_{\text{turn}}, \boldsymbol{\omega}, \boldsymbol{\phi}_{\text{R}}$}
        \EndIf
    \ElsIf{$\text{mode} = 2$ \textbf{and} $|\psi_e| \leq \psi_{\text{th}}^{\text{wp}}$}
        \State $\text{mode} \gets 1$
    \EndIf
\EndProcedure

\Statex

\Procedure{ProcessNearBranch}{$\psi_e, k$}
    \If{$\text{mode} = 1$}
        \If{$|\psi_e| \leq \psi_{\text{th}}^{\text{wp}}$}
            \State $k \gets k + 1$
            \Comment{Waypoint reached}
        \ElsIf{$\psi_e > +\psi_{\text{th}}^{\text{wp}}$}
            \State $\text{mode} \gets 3$
            \State \Call{UpdateCPG}{$\mathbf{a}_{\text{turn}}, \boldsymbol{\omega}, \boldsymbol{\phi}_{\text{L}}$}
        \ElsIf{$\psi_e < -\psi_{\text{th}}^{\text{wp}}$}
            \State $\text{mode} \gets 3$
            \State \Call{UpdateCPG}{$\mathbf{a}_{\text{turn}}, \boldsymbol{\omega}, \boldsymbol{\phi}_{\text{R}}$}
        \EndIf
    \ElsIf{$\text{mode} = 3$ \textbf{and} $|\psi_e| \leq \psi_{\text{th}}^{\text{wp}}$}
        \State $\text{mode} \gets 1$
    \EndIf
\EndProcedure

\Statex

\Function{ModifyAmplitudes}{$\mathbf{a}_0, \delta$}
    \State $\mathbf{a} \gets \mathbf{a}_0$
    \State $a_2 \gets a_2 + 2\delta$, \quad $a_4 \gets a_4 + \delta$
    \State $a_8 \gets a_8 - \delta$, \quad $a_{10} \gets a_{10} - 2\delta$
    \State \Return $\text{clamp}(\mathbf{a}, -70^{\circ}, +70^{\circ})$
\EndFunction

\end{algorithmic}

\vspace{\baselineskip}

Algorithm 1 presents the closed-loop waypoint tracking controller that integrates CPG-based sidewinding with feedback-driven navigation. The controller operates through two distance-based branches: a FAR branch ($>0.35$ m) that prioritizes forward progress with steering corrections, and a NEAR branch ($\leq 0.35$ m) that emphasizes heading alignment before waypoint completion.

\vspace{\baselineskip}


\noindent\textbf{Angle wrapping function:}
\begin{equation}
\text{wrap}(\alpha) = \text{atan2}(\sin\alpha, \cos\alpha)
\end{equation}

\noindent\textbf{Proximity-based blending weight} (trapezoidal with smoothstep):
\begin{equation}
w(d) = 
\begin{cases}
1 & d \leq 0 \\[3pt]
1 - \frac{1}{2} S\!\left(\frac{d}{0.5}\right) & 0 < d \leq 0.5 \\[6pt]
0.5 & 0.5 < d \leq 1.0 \\[3pt]
\frac{1}{2} - \frac{1}{2} S\!\left(\frac{d - 1}{0.5}\right) & 1.0 < d \leq 1.5 \\[6pt]
0 & d > 1.5
\end{cases}
\quad \text{where } S(t) = 3t^2 - 2t^3
\label{eq:weight_function}
\end{equation}
\vspace{0.5em}

\noindent\textbf{Controller parameters:}

\begin{table}[H]
\centering
\small
\begin{tabular}{@{}lccp{6cm}@{}}
\toprule
Parameter & Symbol & Value & Description \\
\midrule
Distance threshold & $d_{\text{th}}$ & 0.35 m & Distance below which robot is "NEAR" waypoint \\
Waypoint yaw threshold & $\psi_{\text{th}}^{\text{wp}}$ & $30^{\circ}$ & Yaw tolerance for waypoint completion \\
Turn trigger threshold & $\psi_{\text{th}}^{\text{turn}}$ & $40^{\circ}$ & Yaw error threshold to trigger turn-in-place \\
Steering offset & $\delta_0$ & $14^{\circ}$ & Base steering offset (nominal sidewind direction) \\
Steering limit & $\delta_{\text{lim}}$ & $7^{\circ}$ & Maximum steering adjustment \\
Proportional gain & $K_p$ & 1.0 & Proportional gain for steering \\
\bottomrule
\end{tabular}
\end{table}

\vspace{\baselineskip}

The controller employs an angle wrapping function to ensure all angular errors remain within the $[-\pi, \pi]$ range, preventing discontinuities when the robot crosses the $\pm 180^\circ$ boundary. This is achieved through the standard $\text{atan2}$ function applied to the sine and cosine of the angle.

\vspace{\baselineskip}

The controller parameters documented in the table were empirically tuned through extensive experimental trials. The distance threshold of 0.35 m defines the transition between far-field navigation and near-field precision alignment, while the yaw thresholds ($30^\circ$ and $40^\circ$) establish clear boundaries for mode switching. The steering parameters ($\delta_0 = 14^\circ$, $\delta_{\text{lim}} = 7^\circ$) were selected to provide sufficient directional control while maintaining the physical constraints of the sidewinding gait pattern.

\begin{figure}[H]
    \centering
    \includegraphics[width=1\linewidth]{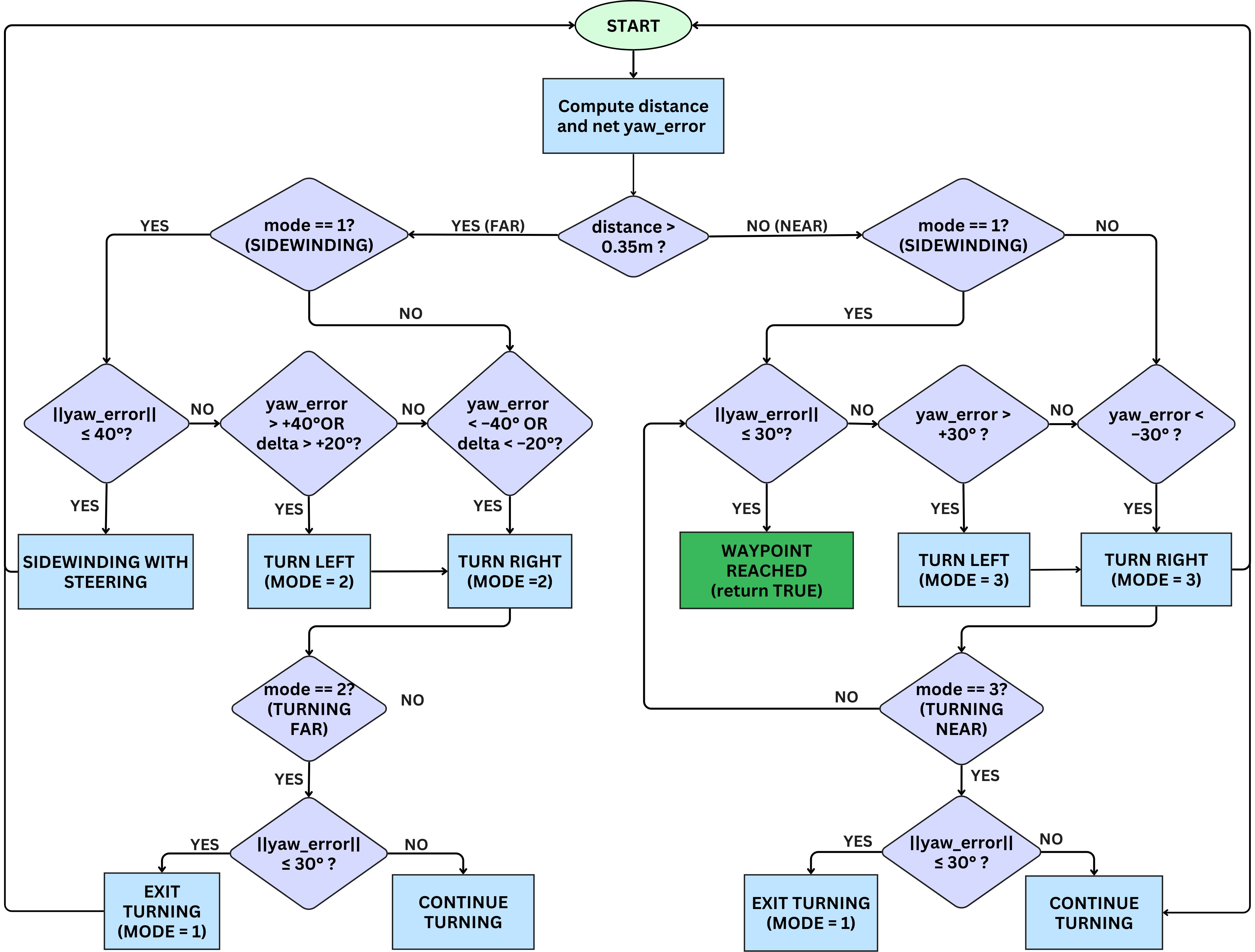}
    \caption{Flowchart representing the controller's logic based on the distance to waypoint and net yaw error.}
    \label{fig:controller_logic_flowchart}
\end{figure}

Figure 3.6 visualizes the controller as a state machine, showing transitions between sidewinding and turn-in-place modes based on distance and yaw error conditions. The flowchart illustrates the hierarchical decision structure where an initial distance check determines the FAR or NEAR branch, followed by mode-specific yaw comparisons. The diagram highlights three operational modes (sidewinding, FAR turning, NEAR turning) with well-defined transition conditions and asymmetric left/right turn handling.

\vspace{\baselineskip}

\begin{table}[H]
\centering
\caption{Action Selection Based on Mode and Errors}
\begin{tabular}{|l|c|l|l|}
\hline
\textbf{Distance} & \textbf{Mode} & \textbf{Yaw Error} & \textbf{Action} \\
\hline
FAR ($>0.35$m) & Sidewind & Small ($\leq 40^\circ$) & Sidewind with steering correction \\
\hline
FAR ($>0.35$m) & Sidewind & Large ($>40^\circ$) & Switch to turn-in-place \\
\hline
FAR ($>0.35$m) & Turning & Reduced ($\leq 30^\circ$) & Switch back to sidewind \\
\hline
NEAR ($\leq 0.35$m) & Sidewind & Small ($\leq 30^\circ$) & \textbf{WAYPOINT REACHED!} \\
\hline
NEAR ($\leq 0.35$m) & Sidewind & Large ($>30^\circ$) & Turn to align heading \\
\hline
NEAR ($\leq 0.35$m) & Turning & Aligned ($\leq 30^\circ$) & Switch to sidewind (then detect reached) \\
\hline
\end{tabular}
\label{tab:action_selection}
\end{table}

\vspace{\baselineskip}

Table 3.1 maps the combinations of distance mode, locomotion mode, and yaw error to the corresponding controller actions, delineating six different navigation scenarios. The table provides an intuitive overview of how distance (0.35 m) and yaw thresholds ($30°$ and $40°$) partition the state space into discrete behavioral regions. The controller parameters table documents the empirically tuned values for all thresholds, gains, and offsets used in implementation.

\chapter{Results}
\label{chap:results}

To validate the efficacy of the proposed autonomous navigation system, a comprehensive experimental campaign was conducted on the COBRA platform. All onboard processing was executed on an Nvidia Jetson Orin Nano, which handled real-time state estimation and control. The evaluation was structured into two primary phases: visual-inertial SLAM validation and closed-loop trajectory tracking performance assessment.

\section{Visual-Inertial State Estimation Validation}

Prior to implementing closed-loop trajectory tracking, a robust onboard perception system was established to provide real-time state feedback during autonomous operation. RTAB-Map (Real-Time Appearance-Based Mapping) was integrated with the Intel RealSense D435i depth camera mounted in the robot's head module, running on the Nvidia Jetson Orin Nano. The visual-inertial odometry pipeline was benchmarked against OptiTrack motion capture ground truth across multiple locomotion modes to characterize drift behavior and identify failure conditions specific to serpentine motion.

These experiments confirmed that while RTAB-Map provided sufficiently accurate short-term pose estimates for feedback control, the system exhibited predictable drift characteristics and failure modes specific to serpentine locomotion. Five primary challenge scenarios were identified:

\begin{itemize}
    \item \textbf{Low-texture environments:} Visual feature extraction became sparse, leading to gradual odometry drift.
    \item \textbf{Close proximity to objects:} The depth camera's minimum range limitations caused partial point cloud dropout.
    \item \textbf{Sudden jerky motion:} Aggressive gait transitions exceeded the IMU's ability to bridge visual frame alignment.
    \item \textbf{Head impact with the ground:} Vertical undulation temporarily occluded the camera's field of view and interrupted tracking.
    \item \textbf{Pure rotation with no translation:} Turn-in-place maneuvers lacked sufficient parallax cues, degrading loop closure detection.
\end{itemize}

\begin{figure}[H]
    \centering
    \includegraphics[width=1\linewidth]{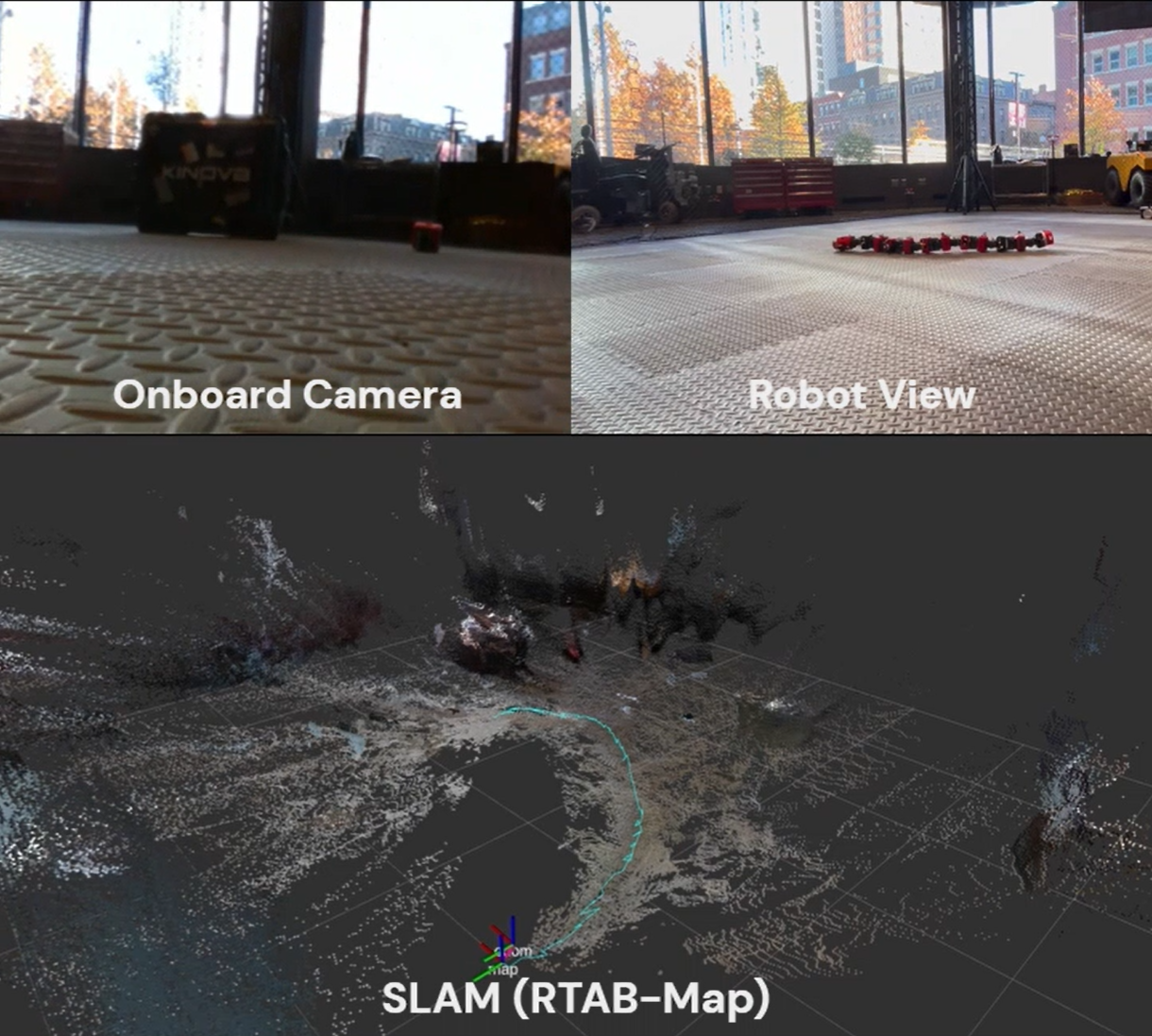}
    \caption{Figure shows SLAM (RTAB-Map) output with the onboard camera view and the external robot view for experiment 1}
    \label{fig:test_setup_1_complete}
\end{figure}

Despite these challenges, the accumulated position error remained bounded within acceptable limits, and the system demonstrated reliable recovery once the sidewinding motion resumed in certain cases, making it suitable as the foundation for closed-loop trajectory tracking.

Fig.~\ref{fig:test_setup_1_complete} presents representative snapshots from the validation experiment, showing onboard camera views paired with RTAB-Map visual-inertial odometry and generated point cloud map. The robot performed curved sidewinding locomotion in the high-bay drone cage while executing SLAM onboard.

\begin{figure}[H]
    \centering
    \includegraphics[width=1\linewidth]{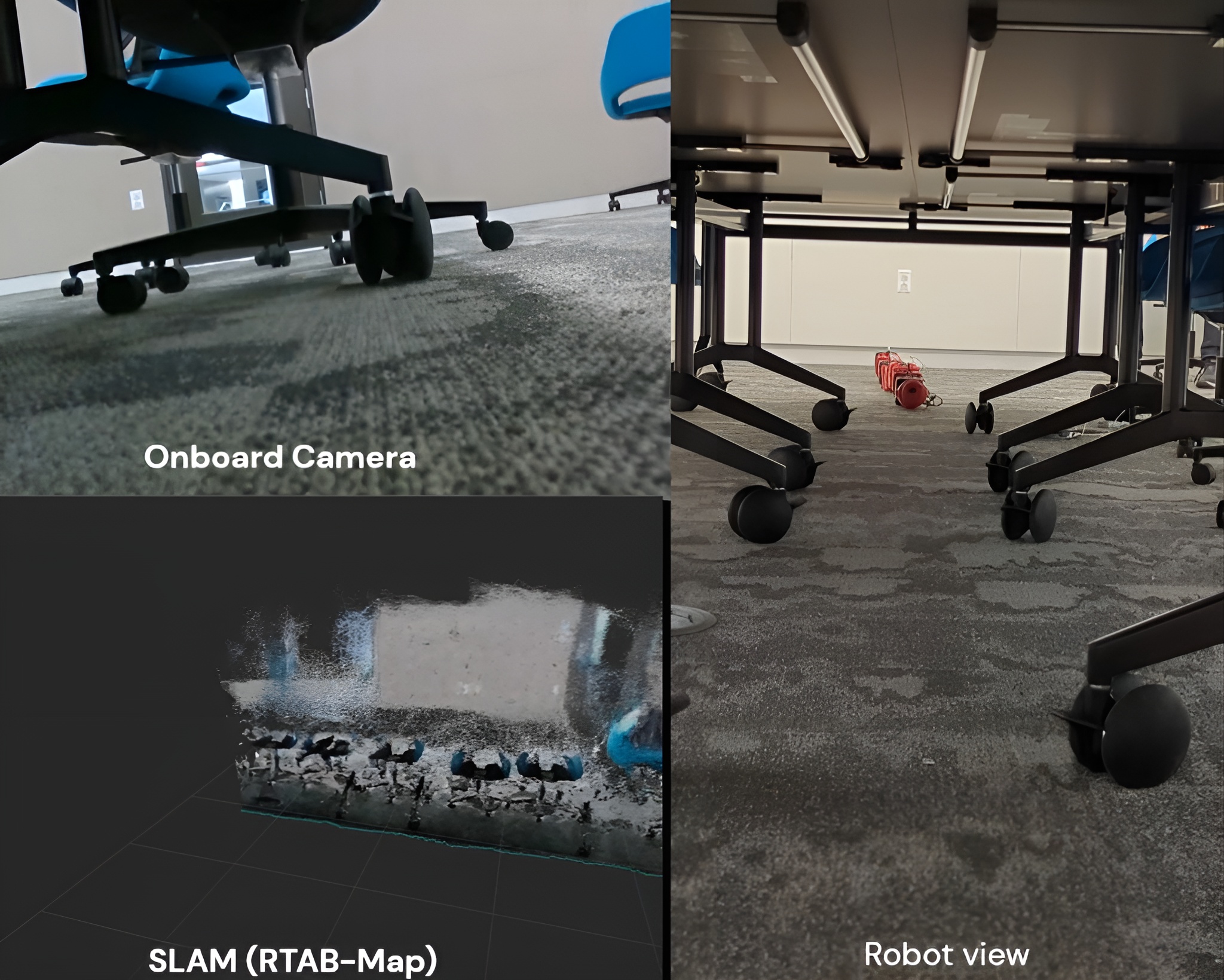}
    \caption{Figure shows SLAM (RTAB-Map) output with the onboard camera view and the external robot view for experiment 2}
    \label{fig:test_setup_2_complete}
\end{figure}

Fig.~\ref{fig:test_setup_2_complete} demonstrates SLAM performance during vertical undulation (worm gait) in the EXP conference room. The robot executed straight-line locomotion through an environment filled with visually identical chairs. Despite the repetitive features that could confuse loop closure detection, RTAB-Map maintained stable tracking, benefiting from the worm gait's slower velocity which reduced motion blur and improved visual feature correspondence.

\begin{figure}[H]
    \centering
    \includegraphics[width=1\linewidth]{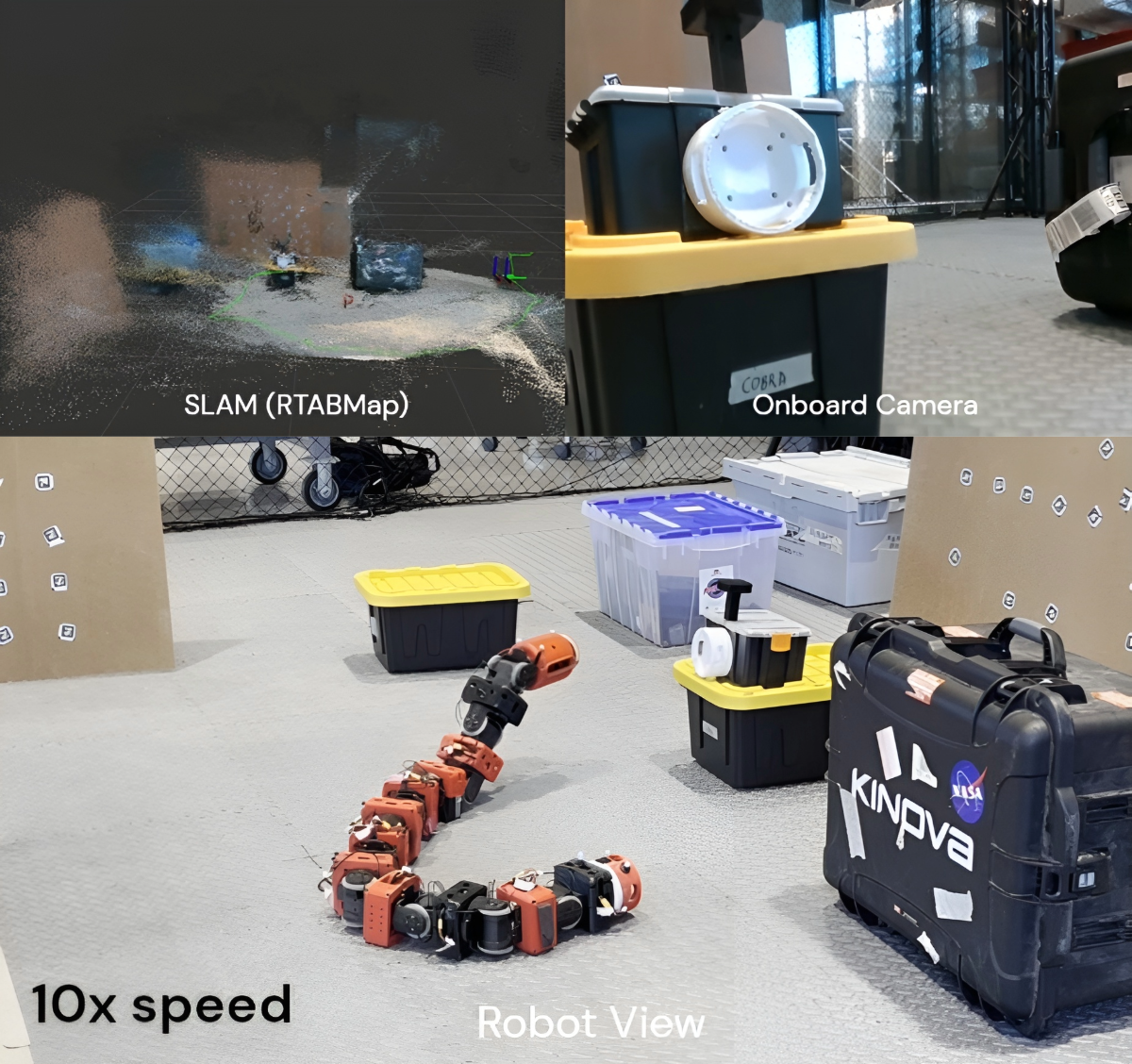}
    \caption{Figure shows SLAM (RTAB-Map) output with the onboard camera view and the external robot view for experiment 3}
    \label{fig:test_setup_3_complete}
\end{figure}

Fig.~\ref{fig:test_setup_3_complete} shows SLAM performance during navigation and inspection in a cluttered high-bay environment.The robot traversed toward a target box surrounded by cardboard sheets and obstacles, then executed a survey maneuver to inspect the object and its surroundings. Despite the varied textures, colors, and geometric features, RTAB-Map maintained stable tracking throughout, demonstrating robust performance across diverse visual scene content.

\begin{figure}[H]
    \centering
    \includegraphics[width=1\linewidth]{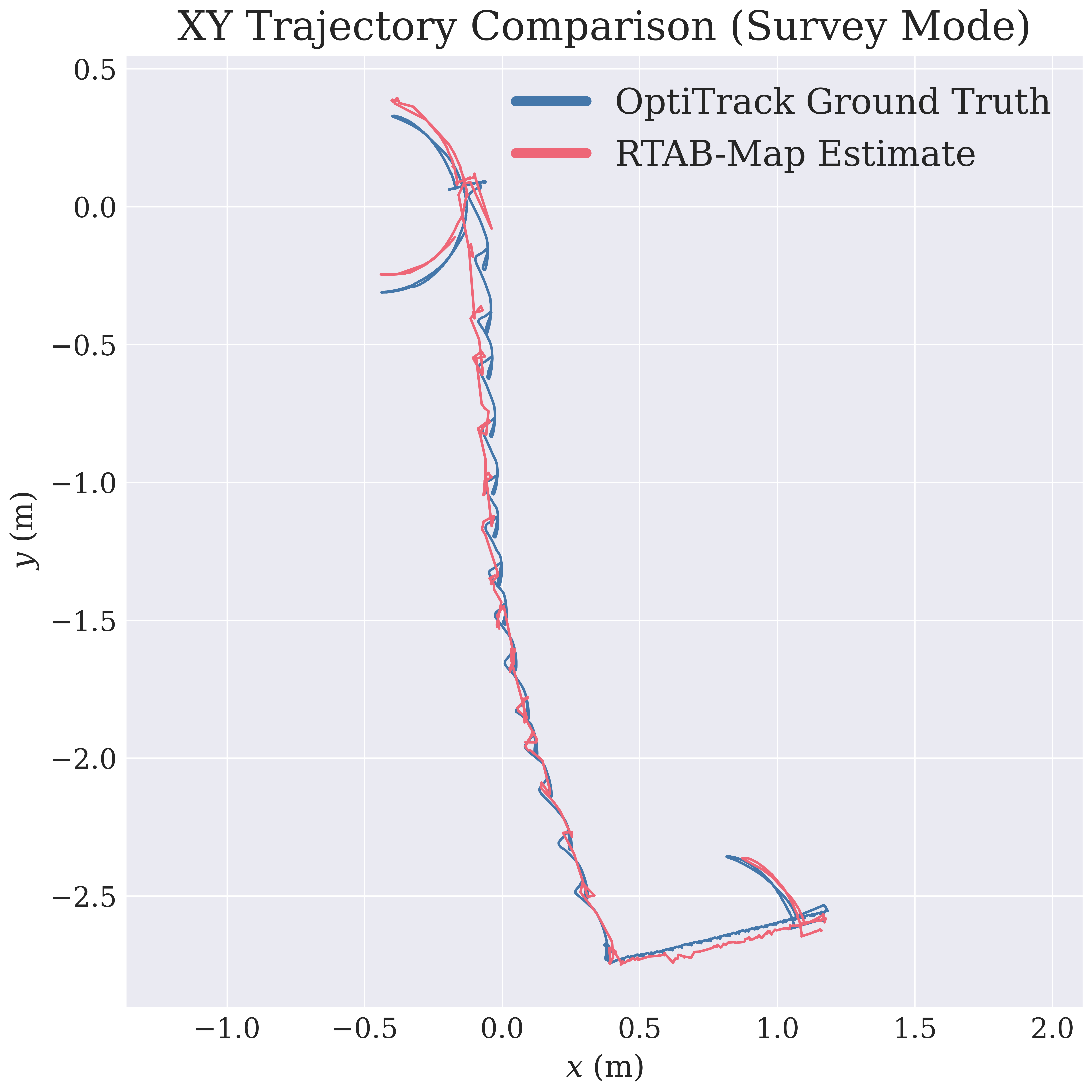}
    \caption{Illustrates the comparison between XY trajectories of RTAB-Map estimate and OptiTrack Ground Truth for experiment 2}
    \label{fig:optitrack_good_XY__comparison}
\end{figure}

\begin{figure}[H]
    \centering
    \includegraphics[width=1\linewidth]{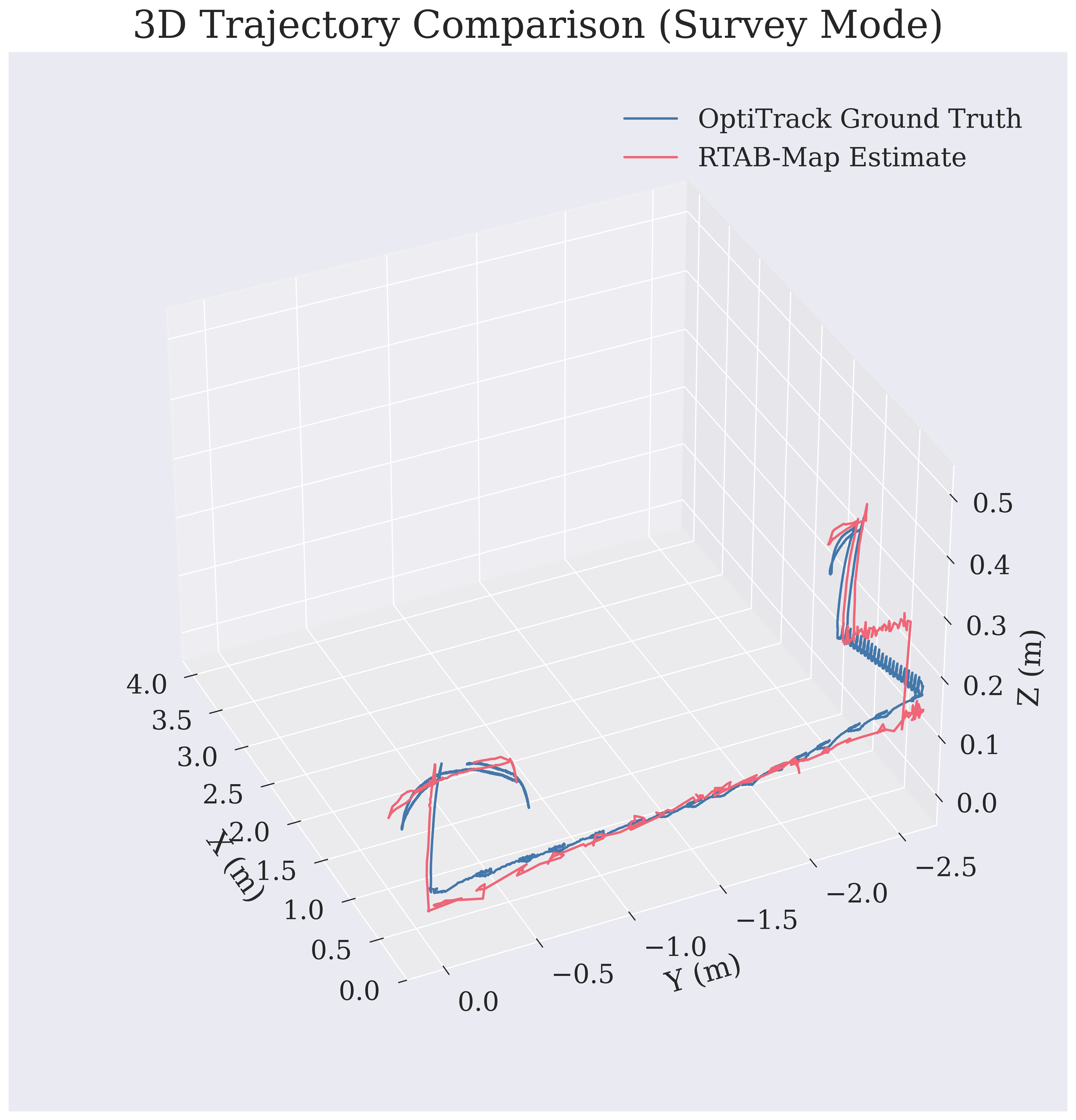}
    \caption{Illustrates the comparison between XYZ trajectories of RTAB-Map estimate and OptiTrack Ground Truth for experiment 2}
    \label{fig:optitrack_good_XYZ_comparison}
\end{figure}

Fig.~\ref{fig:optitrack_good_XY__comparison} and Fig.~\ref{fig:optitrack_good_XYZ_comparison} compares the estimated trajectory against OptiTrack ground truth for experiment shown in Fig.~\ref{fig:test_setup_3_complete}. Fig.~\ref{fig:optitrack_good_XY__comparison} overlays the XY trajectories, while Fig.~\ref{fig:optitrack_good_XYZ_comparison} plots X, Y, and Z position errors over time. The system achieved a maximum absolute pose error of 13.41 cm, mean error of 6.85 cm, and RMSE of 7.33 cm, demonstrating that RTAB-Map provided sufficiently accurate state estimation.

\begin{figure}[H]
    \centering
    \includegraphics[width=1\linewidth]{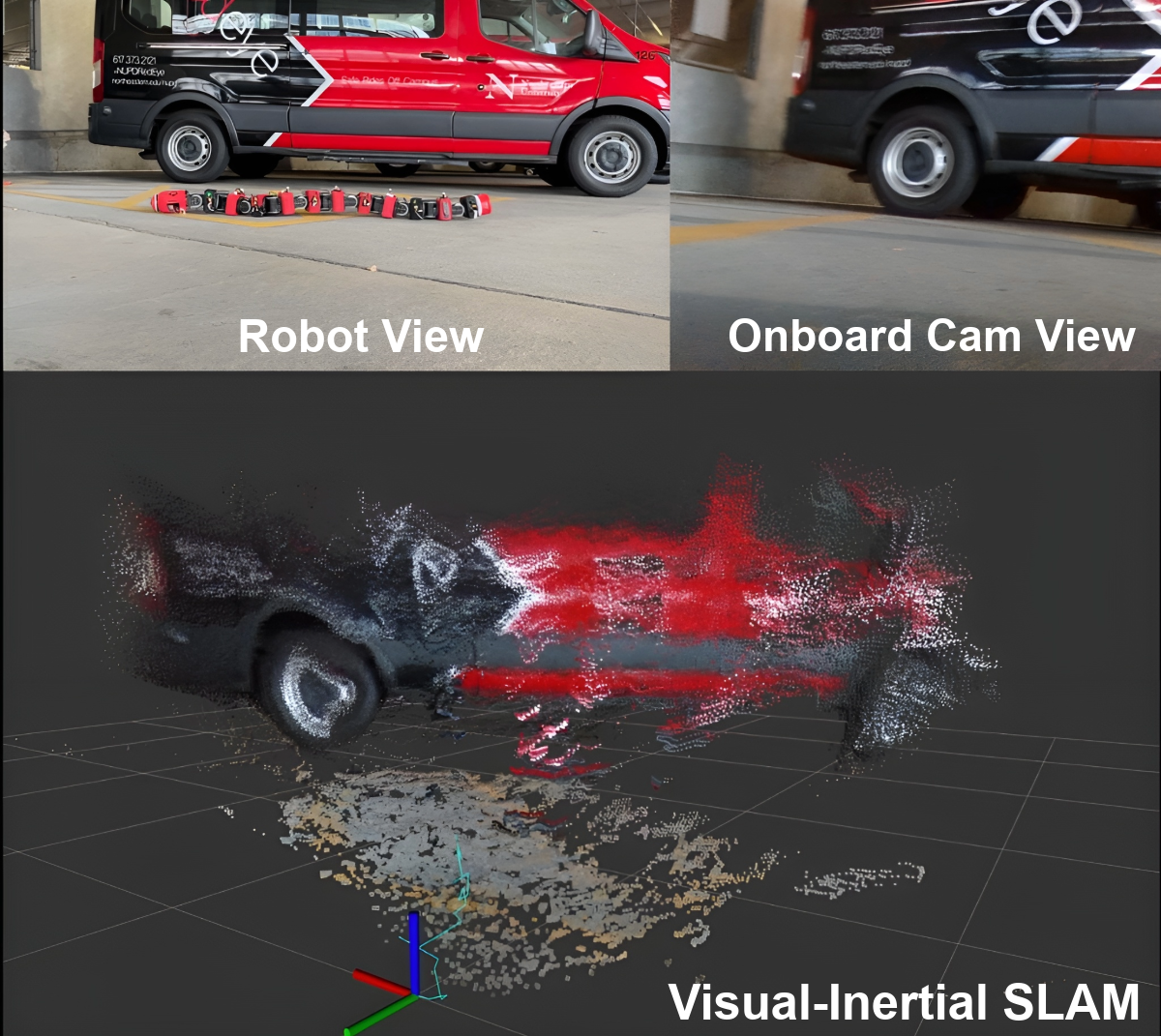}
    \caption{Figure shows SLAM (RTAB-Map) output with the onboard camera view and the external robot view for experiment 4}
    \label{fig:test_setup_5_complete}
\end{figure}

Fig.~\ref{fig:test_setup_5_complete} illustrates a failure case under extreme conditions. The robot attempted to navigate beneath parked vans in a garage on high-friction asphalt. RTAB-Map tracked successfully during the approach but failed immediately upon entering the shadowed region under the vehicle, where lighting was insufficient for visual feature detection. This experiment confirmed that active illumination would be necessary for reliable state estimation in low-light environments such as crater shadows or subsurface cavities.

\section{Closed-Loop Trajectory Tracking Performance}

With validated state estimation established, forward kinematics processed joint encoder data from all 11 Dynamixel servos alongside the head pose to compute the Center of Mass pose at 100 Hz, providing the feedback signal for trajectory tracking. An OptiTrack motion capture system provided ground-truth head pose at 100 Hz, while the high-level feedback control loop operated at 1 Hz.

The trajectory tracking validation was structured into four scenarios: single-waypoint convergence analysis, complex path following (Star pattern), bidirectional repeatability testing (Home retreat), and disturbance rejection. The following sections detail the quantitative performance of the system in each scenario.

\subsection{Single-Waypoint Convergence Analysis}

The first set of experiments evaluated the controller’s ability to regulate the robot to a specific setpoint from arbitrary initial conditions. A single waypoint was defined at the origin of the world frame $(0, 0)$ with a target yaw of $0^\circ$. The robot was initialized at 13 different configurations across the arena, varying in both Euclidean distance from the origin and initial orientation.

\begin{figure}[H]
    \centering
    \includegraphics[width=1\linewidth]{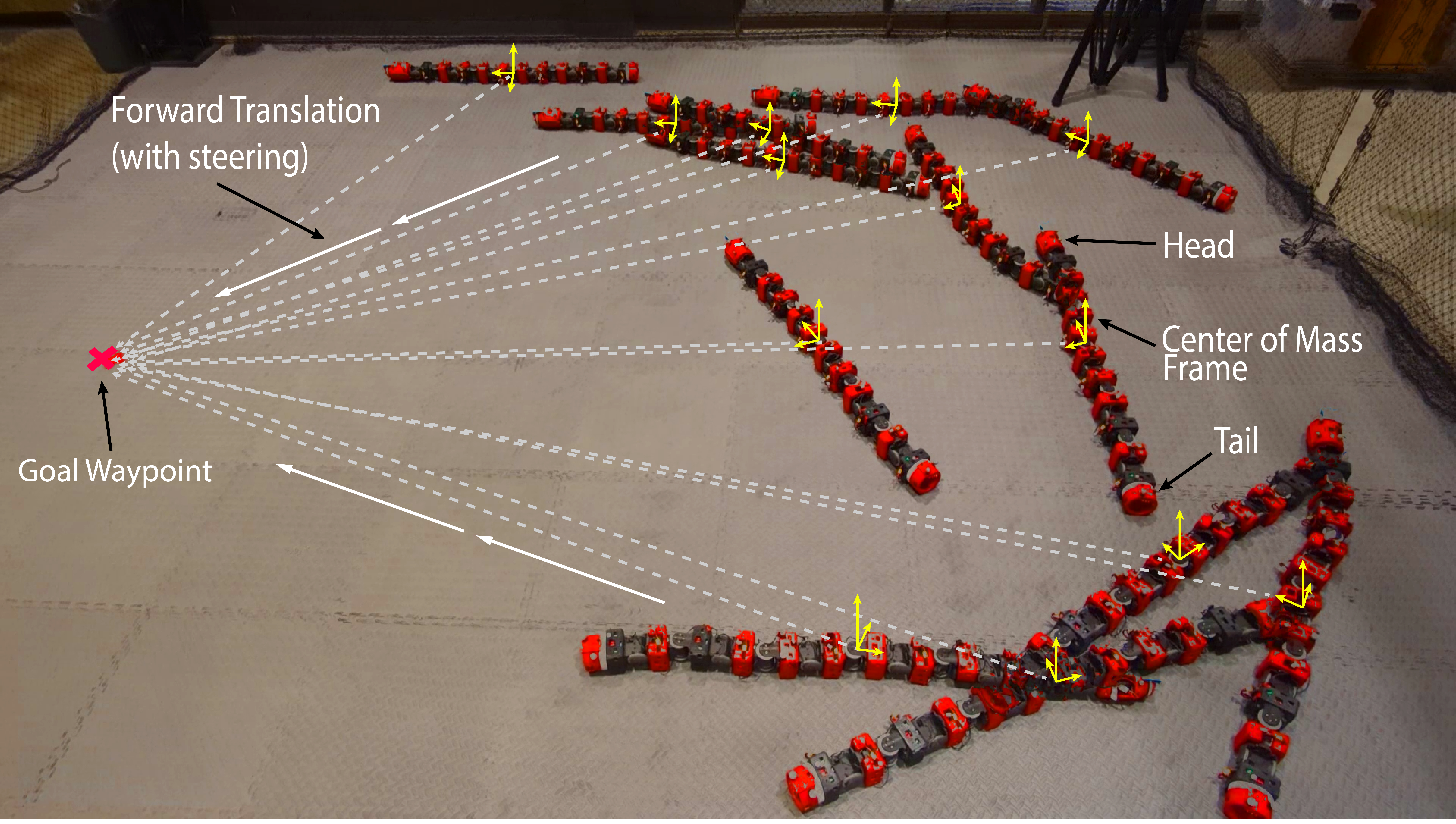}
    \caption{Shows all initial configurations prior to movement toward the waypoint.}
    \label{fig:single_waypoint_convergence_frame_1}
\end{figure}

\begin{figure}[t!]
    \centering
    \includegraphics[width=1\linewidth]{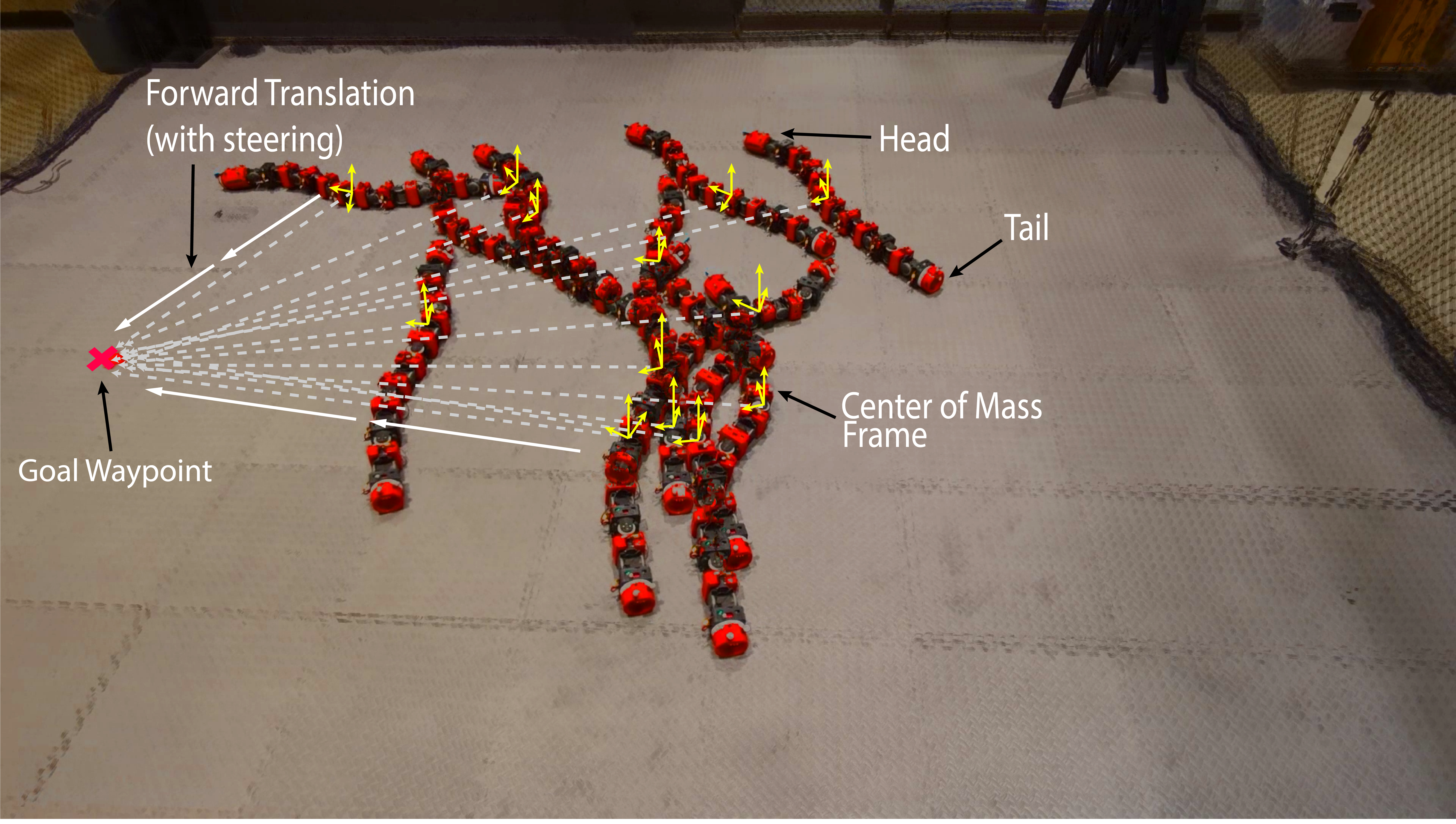}
    \caption{Shows all intermediate configurations during their progression toward the waypoint.}
    \label{fig:single_waypoint_convergence_frame_2}
\end{figure}

\begin{figure}[t!]
    \centering
    \includegraphics[width=1\linewidth]{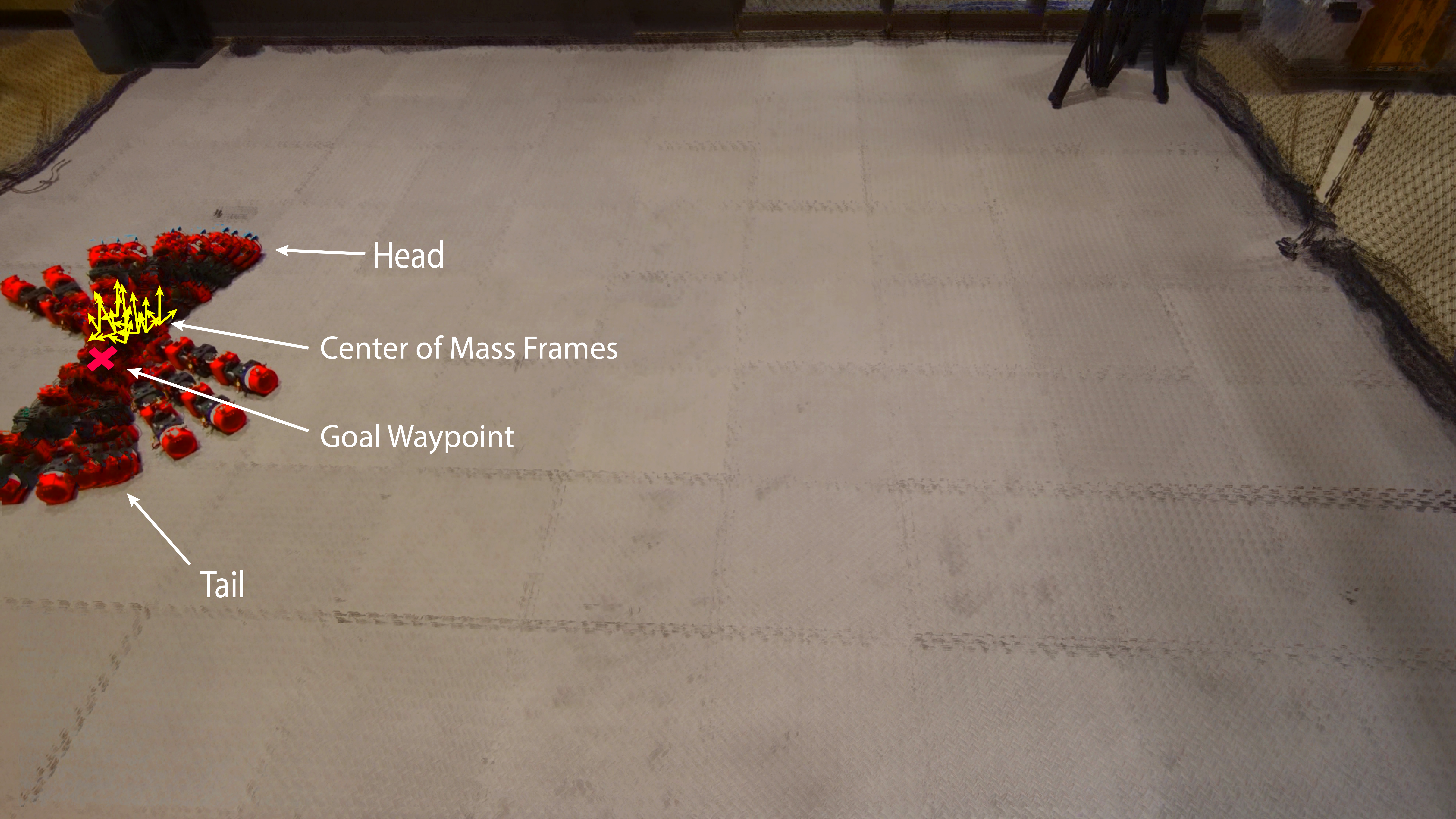}
    \caption{Shows all final configurations upon converging at the waypoint.}
    \label{fig:single_waypoint_convergence_frame_3}
\end{figure}

\begin{figure}[t!]
    \centering
    \includegraphics[width=1\linewidth]{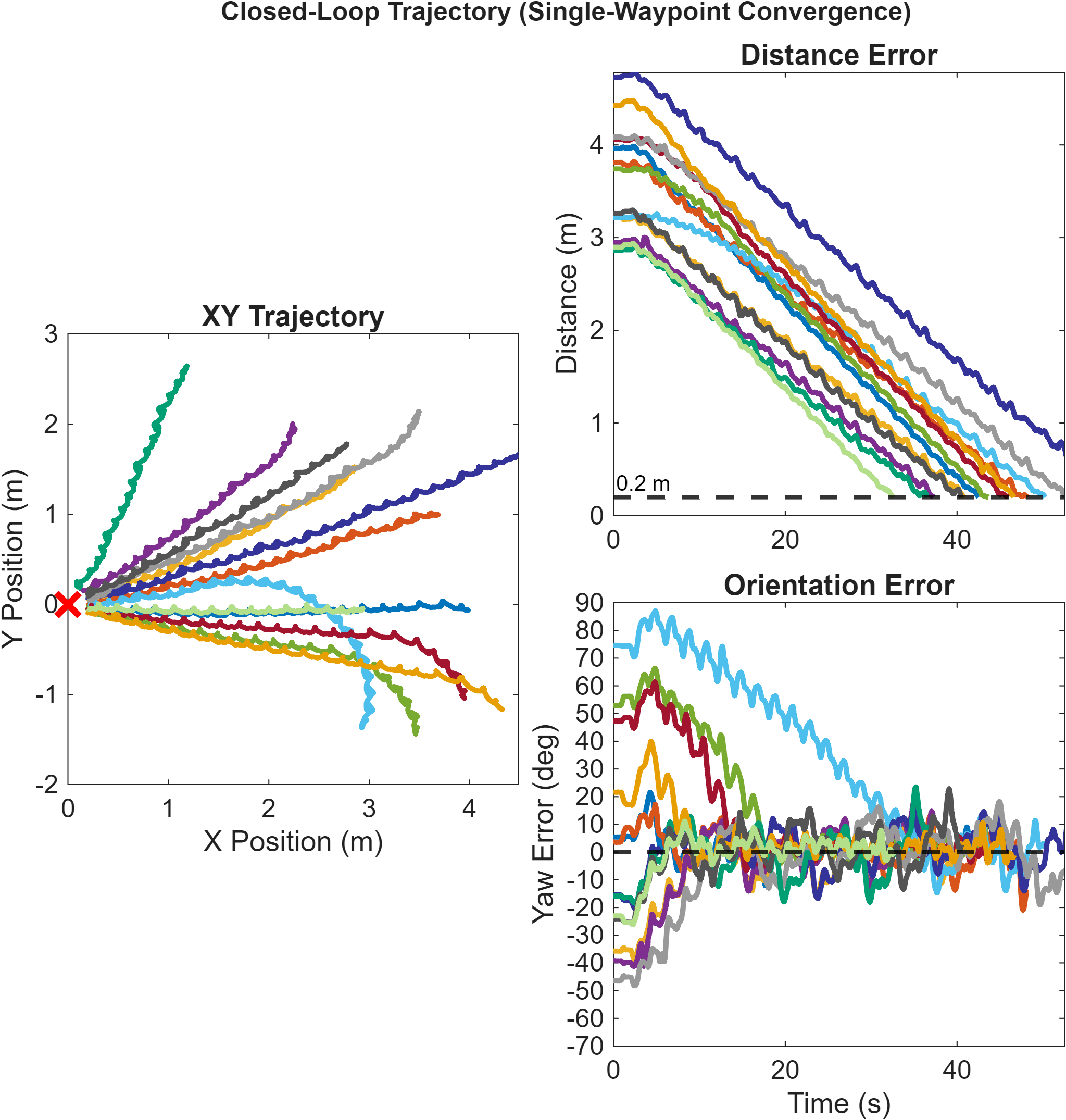}
    \caption{Illustrates the XY trajectory and the distance and orientation error with respect to time for the single waypoint convergence experiment}
    \label{fig:single_waypoint_convergence_analysis}
\end{figure}

Fig.~\ref{fig:single_waypoint_convergence_analysis} presents the composite results of these 13 trials. (a) illustrates the XY trajectories of the robot’s Center of Mass (CoM). The trajectories demonstrate a consistent convergence behavior; regardless of the starting quadrant or initial heading, the controller successfully guided the robot into a smooth approach vector toward the origin and does not suffer from local minima within the tested workspace.

Subfigure (b) plots the Euclidean distance error over time for all trials. The data shows a monotonic decrease in error, reflecting the stable approach characteristic of the sidewinding gait. The system consistently converges to a steady-state error of approximately 0.2 m, corresponding to the defined stopping threshold ($d_{thresh}$), below which the robot is commanded to halt to avoid limit-cycle oscillations near the waypoint. Simultaneously, subfigure (c) depicts the yaw error convergence. The controller successfully regulates the orientation error to $0^\circ$ in all cases. Notably, the yaw error often decays faster than the distance error, indicating that the robot prioritizes heading alignment during the approach phase to maximize the efficiency of the sidewinding motion.

\subsection{Validation of Complex Trajectory Tracking via Star Topology}

Following the convergence validation, the controller was tested on a multi-segment trajectory defined by a “Star” pattern composed of five waypoints. This required the robot to perform acute turns and sequential waypoint transitions.

\begin{figure}[H]
    \centering
    \includegraphics[width=1\linewidth]{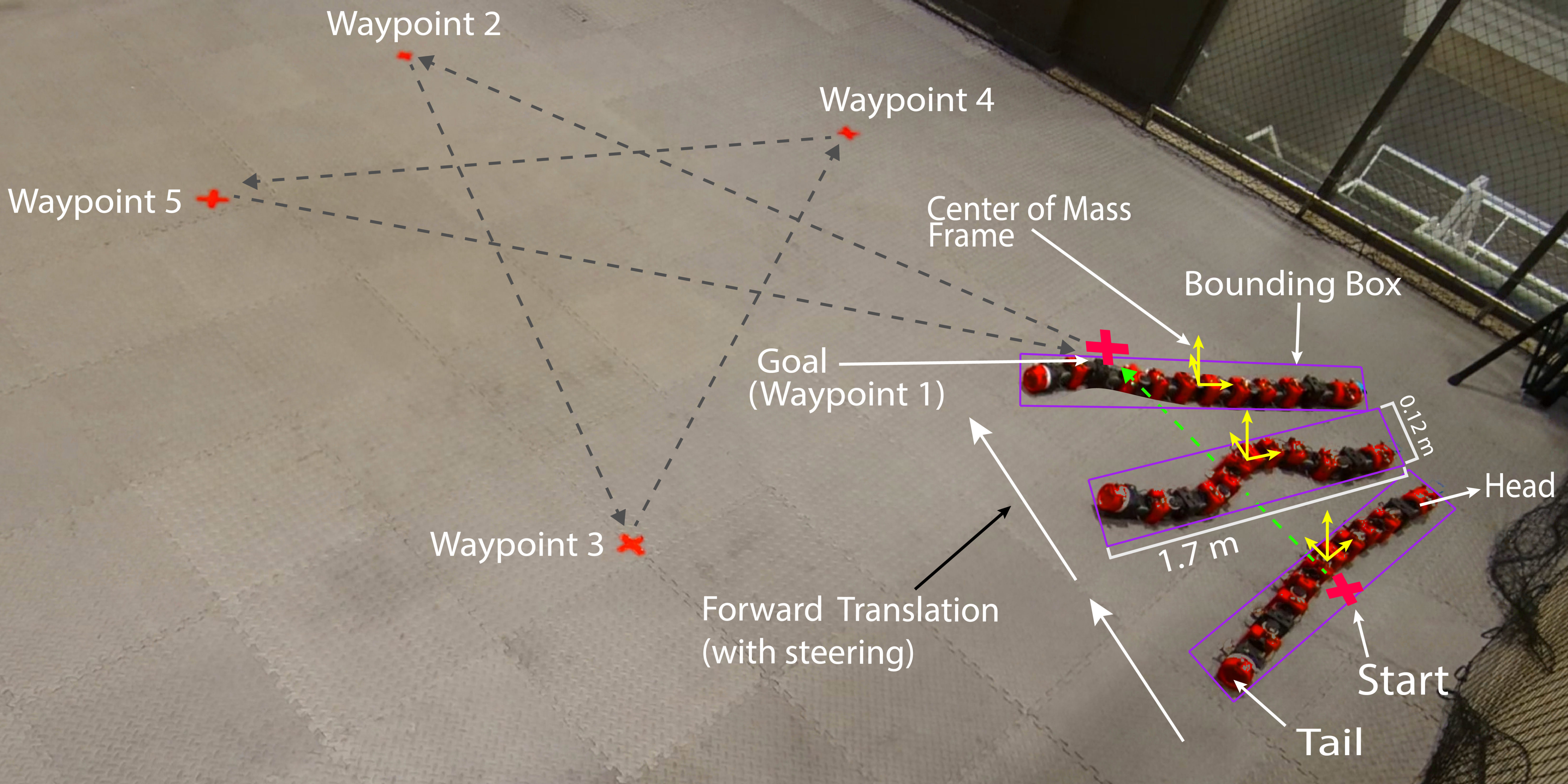}
    \caption{Shows the robot's trajectory from the start position to Waypoint 1}
    \label{fig:star_pattern_closed_loop_frame_1}
\end{figure}

\begin{figure}[H]
    \centering
    \includegraphics[width=1\linewidth]{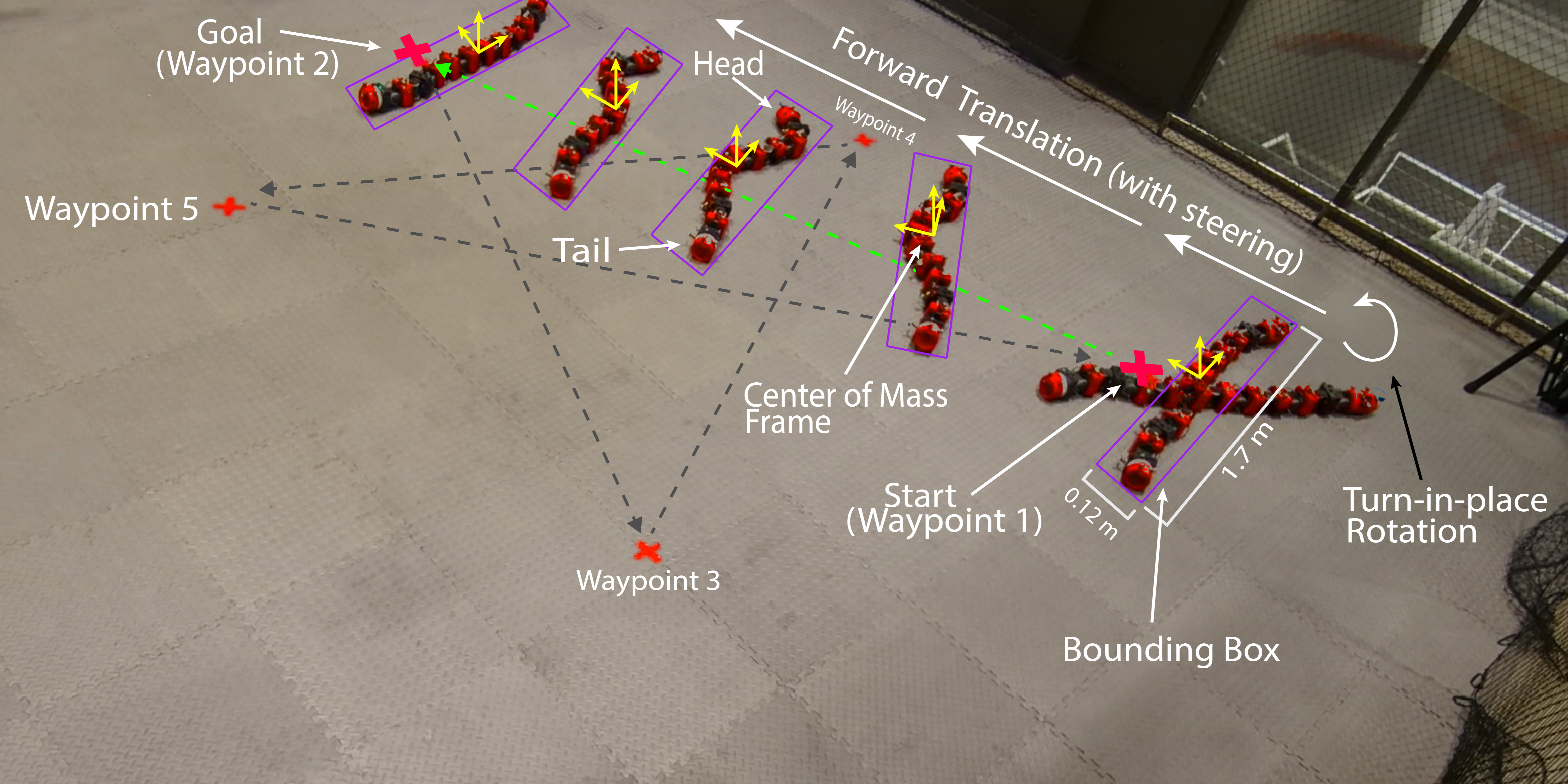}
    \caption{Shows the robot’s trajectory from Waypoint 1 to Waypoint 2}
    \label{fig:star_pattern_closed_loop_frame_2}
\end{figure}

\begin{figure}[H]
    \centering
    \includegraphics[width=1\linewidth]{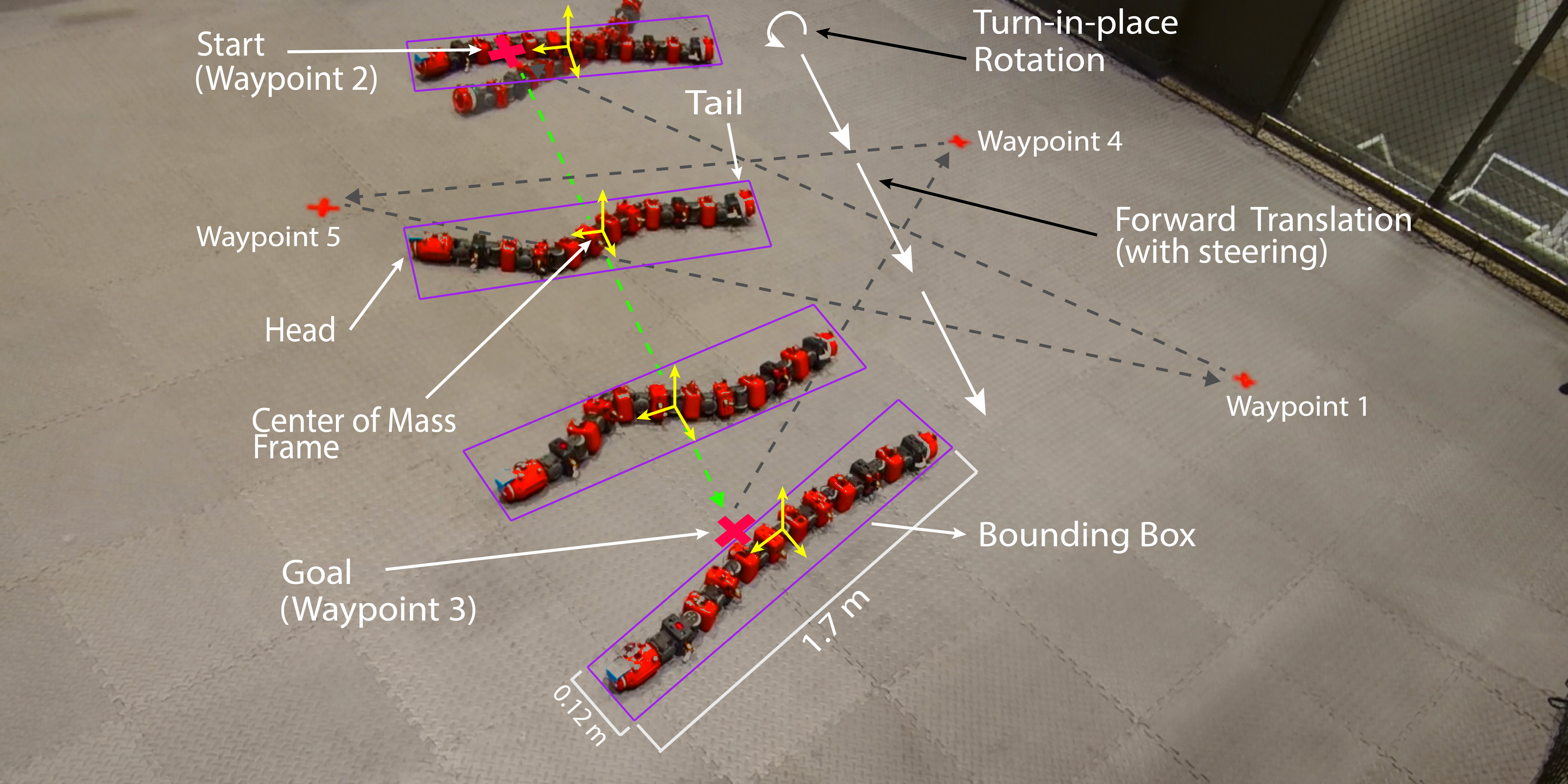}
    \caption{Shows the robot’s trajectory from Waypoint 2 to Waypoint 3}
    \label{fig:star_pattern_closed_loop_frame_3}
\end{figure}

\begin{figure}[H]
    \centering
    \includegraphics[width=1\linewidth]{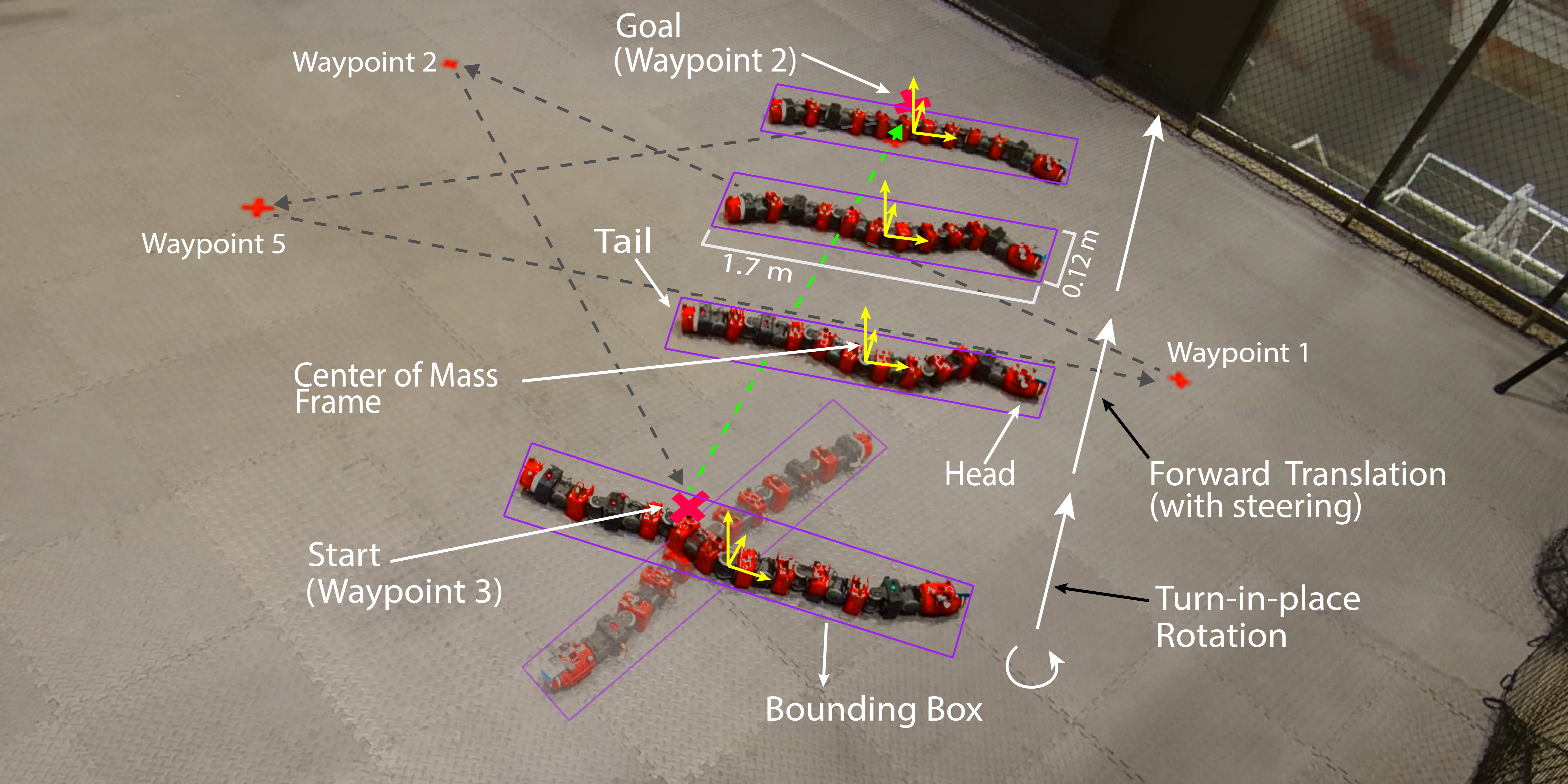}
    \caption{Shows the robot’s trajectory from Waypoint 3 to Waypoint 4}
    \label{fig:star_pattern_closed_loop_frame_4}
\end{figure}

\begin{figure}[H]
    \centering
    \includegraphics[width=1\linewidth]{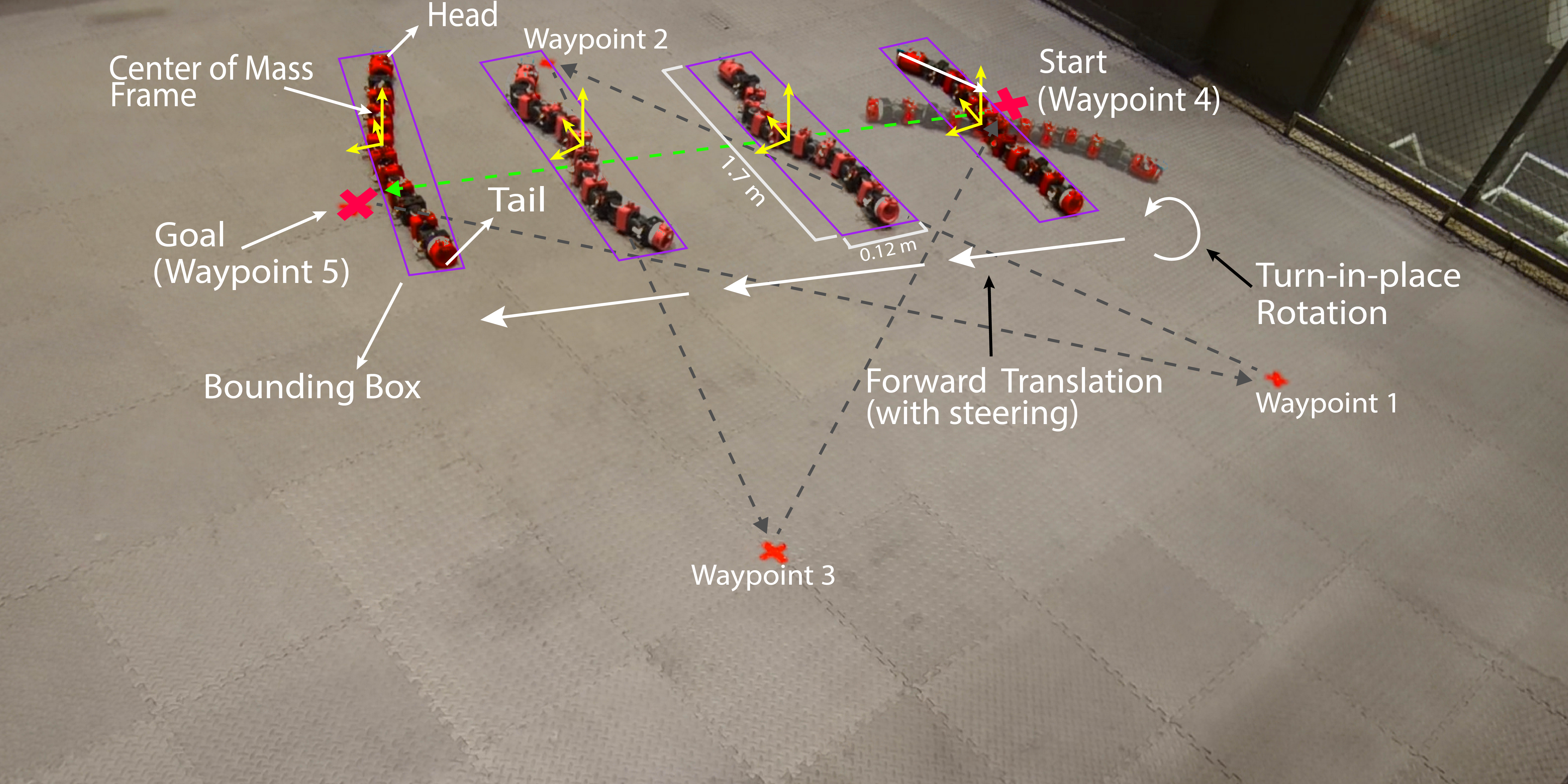}
    \caption{Shows the robot’s trajectory from Waypoint 4 to Waypoint 5}
    \label{fig:star_pattern_closed_loop_frame_5}
\end{figure}

Between Fig.~\ref{fig:star_pattern_closed_loop_frame_1} and Fig.~\ref{fig:star_pattern_closed_loop_frame_5} we present a time-ordered sequence of five snapshots showing the robot’s journey through the commanded waypoint sequence from waypoint 1 to waypoint 5. At each waypoint, the robot performs a brief in-place rotation without lateral translation, after which it begins moving towards the next goal using forward sidewinding while steering to maintain its course. These snapshots illustrate the stepwise reduction in Euclidean distance as well as both relative and absolute yaw error as the robot approaches each successive waypoint.

\begin{figure}[H]
    \centering
    \includegraphics[width=1\linewidth]{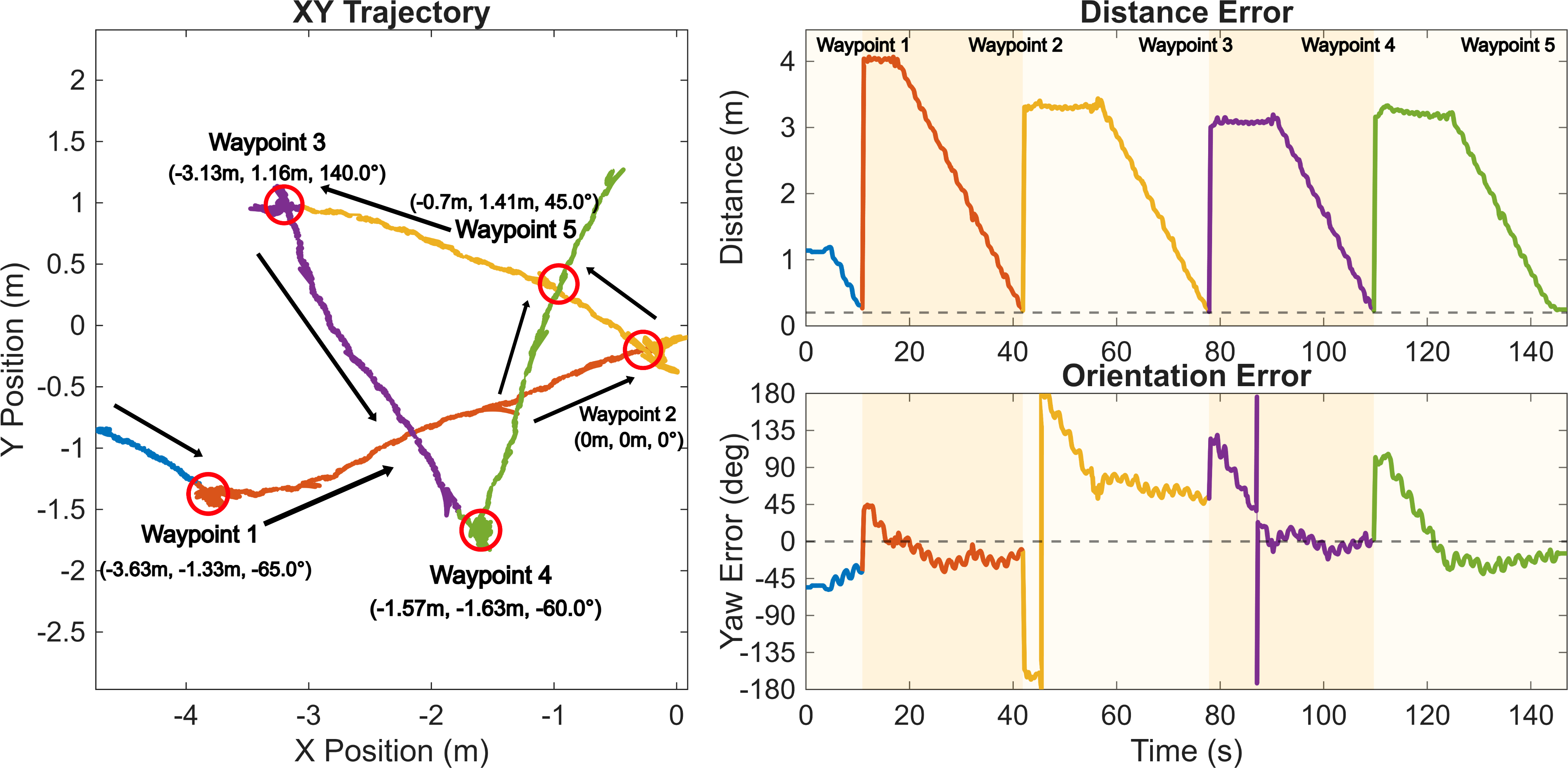}
    \caption{Illustrates the XY trajectory and the distance and orientation error with respect to time for the star topology experiment}
    \label{fig:star_pattern_closed_loop}
\end{figure}

Fig.~\ref{fig:star_pattern_closed_loop} summarizes the tracking performance. Subfigure (a) shows the CoM trajectory overlaid on the desired star vertices, confirming that the robot reached all five waypoints in the correct order. Subfigures (b) and (c) present the distance and orientation errors, with alternating background shading indicating waypoint transitions. The distance error exhibits the expected “sawtooth’’ behavior—decaying toward the 0.2 m threshold and spiking upon switching to the next waypoint. The consistent decay slopes across segments indicate that controller performance is direction-independent within the arena.

\begin{figure}[H]
    \centering
    \includegraphics[width=1\linewidth]{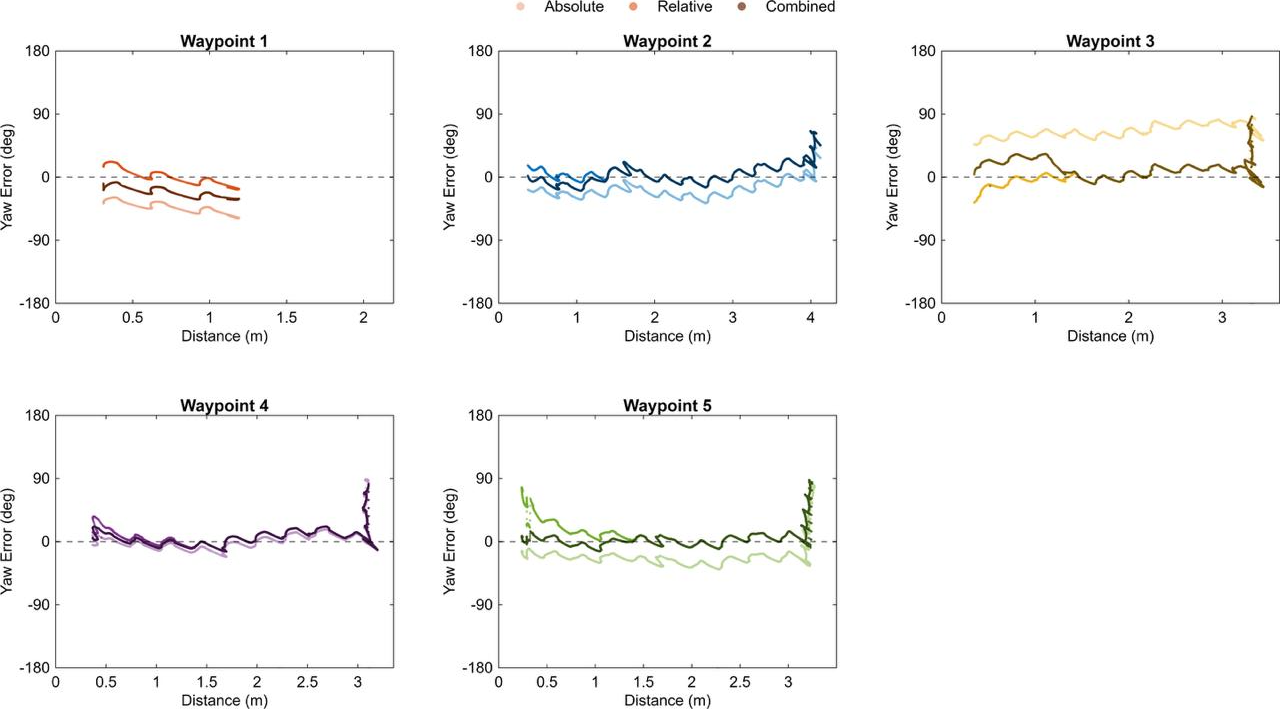}
    \caption{Illustrates the combined yaw error with respect to distance from waypoint for the star topology experiment}
    \label{fig:star_pattern_closed_loop_yaw_vs_distance}
\end{figure}

\begin{figure}[t!]
    \centering
    \includegraphics[width=1\linewidth]{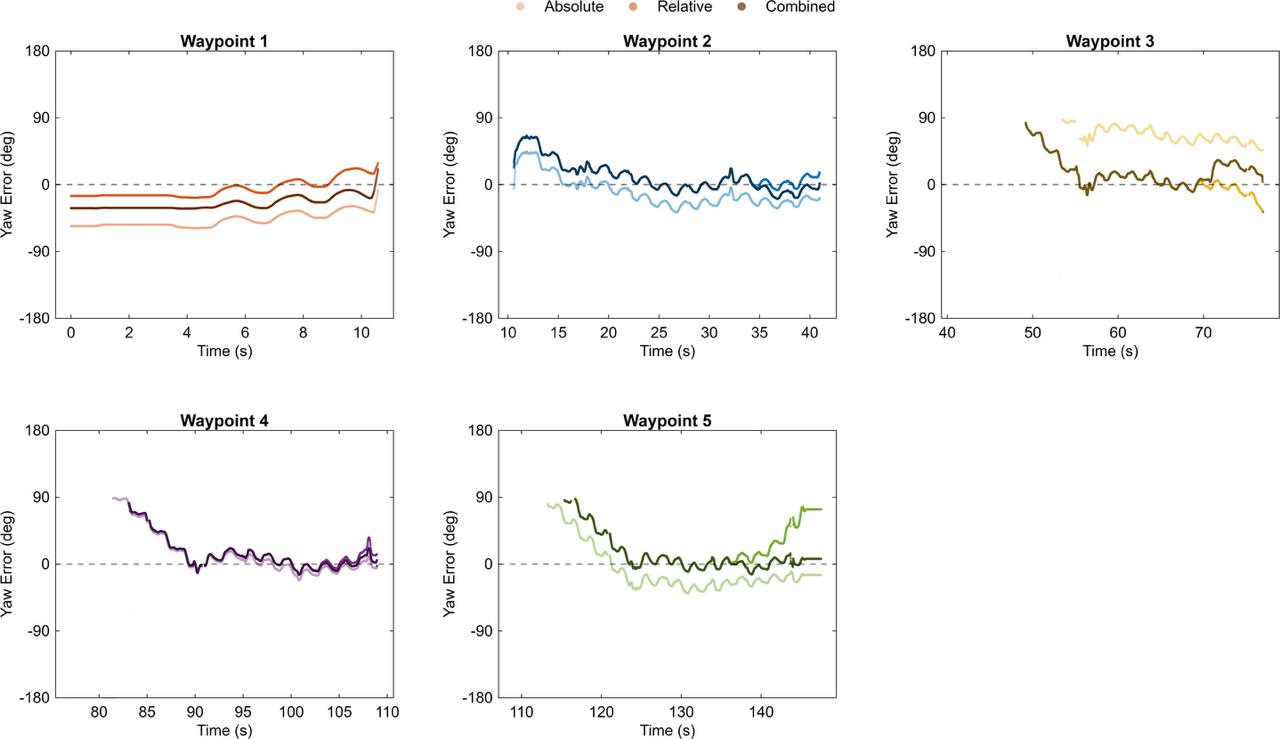}
    \caption{Illustrates the combined yaw error with respect to time for the star topology experiment}
    \label{fig:star_pattern_closed_loop_yaw_vs_time}
\end{figure}

To examine turning behavior more closely, Fig.~\ref{fig:star_pattern_closed_loop_yaw_vs_distance} and Fig.~\ref{fig:star_pattern_closed_loop_yaw_vs_time} plot the yaw error components for each of the five waypoints as functions of distance from the waypoint and time, respectively. Each subplot visualizes the absolute, relative, and combined yaw errors. The relative yaw error decreases rapidly at large distances and during the initial phase of each segment, showing that the robot aligns with the path direction early ("look-then-move" behavior). As the robot nears each vertex, the controller shifts emphasis to absolute yaw, ensuring arrival with the correct final orientation.

\pagebreak

\subsection{Assessment of Spatial Consistency of Bidirectional Trajectories}

To verify how consistent the robot's locomotion is when traveling in opposite directions, we set up an experiment where the robot moved between a central ``Home'' point and four outer targets. The robot visited a target and immediately returned to the center, repeating this four times. This allowed us to directly compare the outbound paths against the return paths across an 8-segment mission.

\begin{figure}[t!]
    \centering
    \includegraphics[width=1\linewidth]{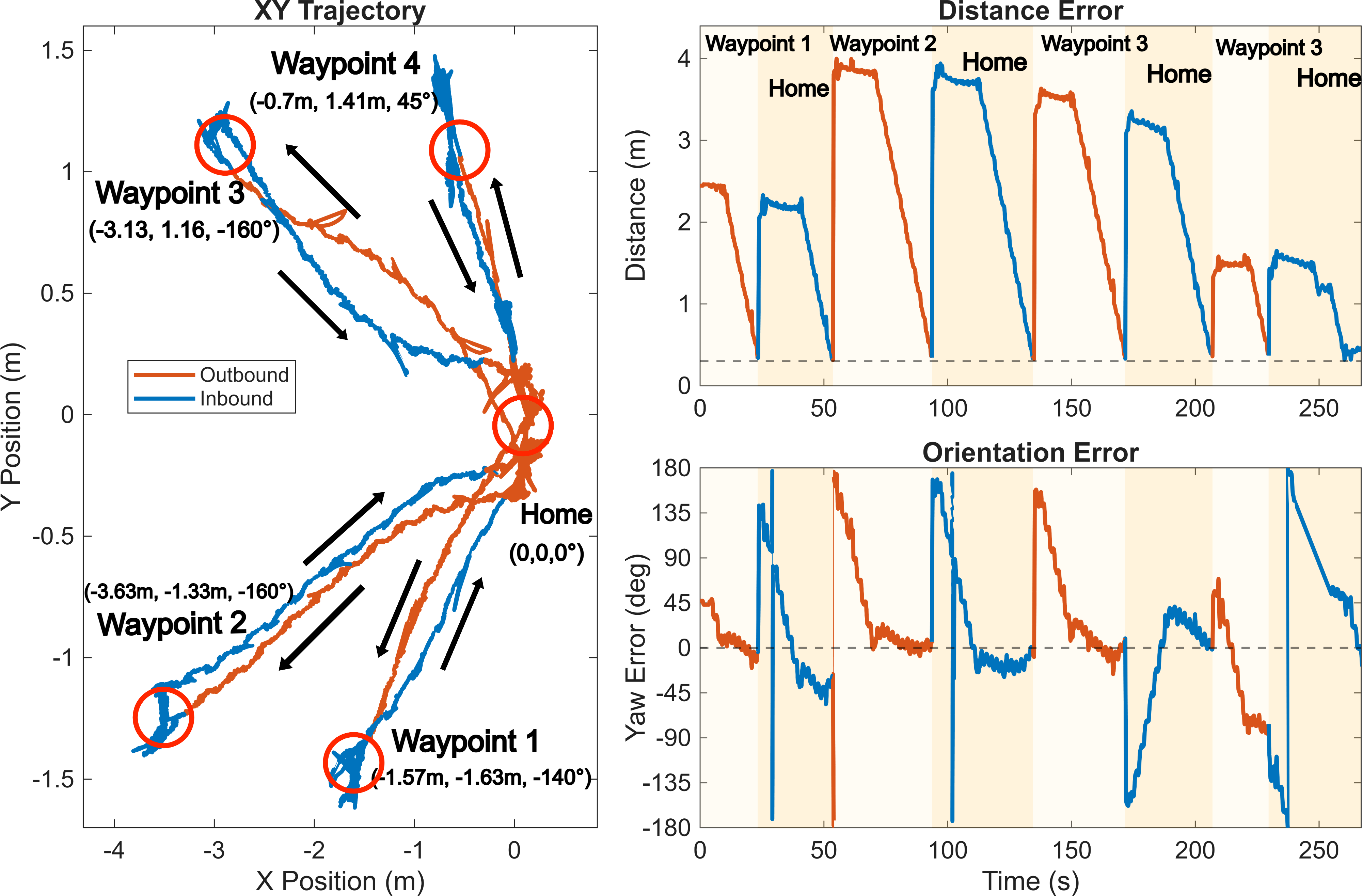}
    \caption{Illustrates the XY trajectory and the distance and orientation error with respect to time for the bidirectional trajectory tracking experiment}
    \label{fig:star_pattern_come_back_origin}
\end{figure}

Fig.~\ref{fig:star_pattern_come_back_origin} summarizes the tracking performance for the bidirectional test. (a) shows the CoM trajectories for the four outbound and four return trips; the return paths naturally mirror the outbound traces and cluster tightly around the central "Home" waypoint, confirming high spatial consistency. (b) and (c) plot the Euclidean distance and orientation errors, respectively, with alternating background shading distinguishing the mission segments. The distance error in (b) follows a distinct sawtooth pattern, spiking when the target switches and decaying as the robot approaches the goal. Crucially, the decay slopes remain consistent across both outbound and return legs, demonstrating that the controller’s convergence performance is direction-independent.

\begin{figure}[H]
    \centering
    \includegraphics[width=1\linewidth]{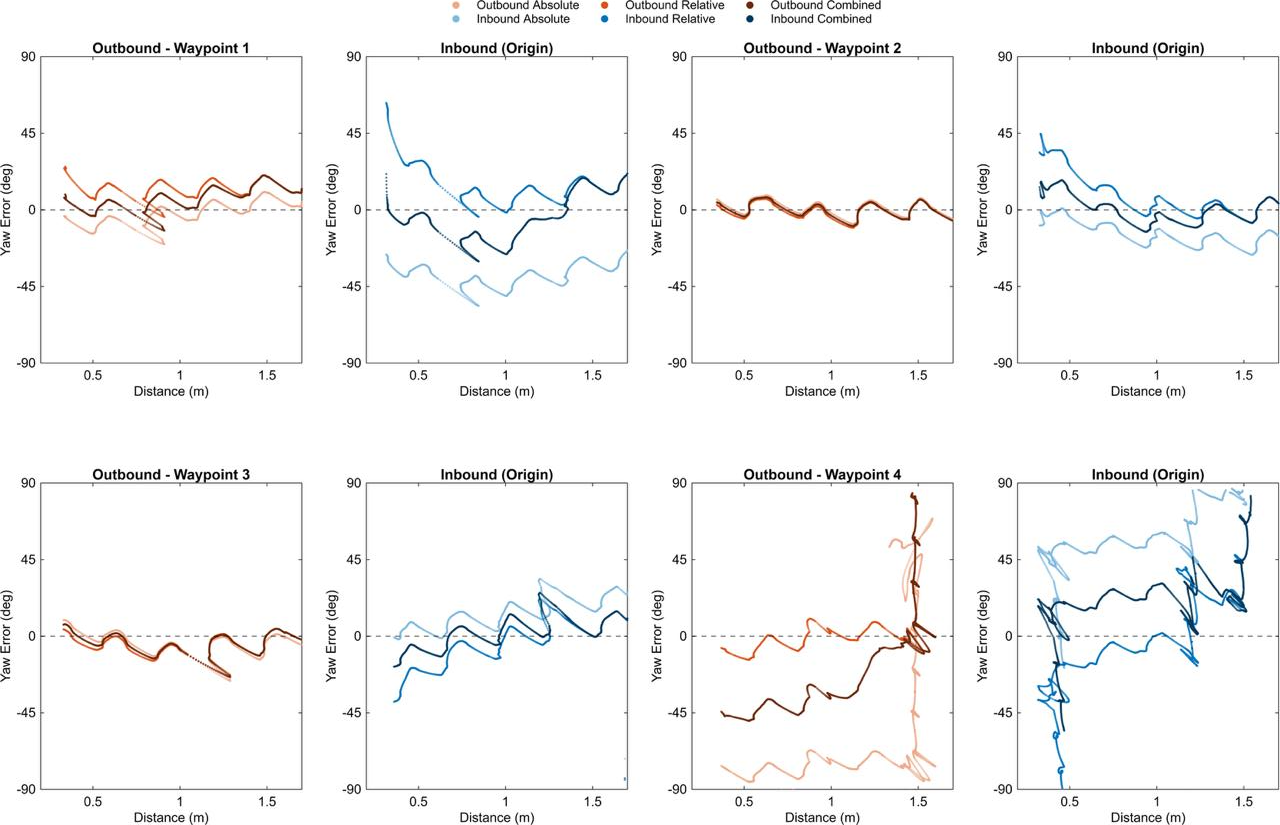}
    \caption{Illustrates the combined yaw error with respect to distance from waypoint for the bidirectional trajectory tracking experiment}
    \label{fig:star_pattern_come_back_origin_yaw_plots}
\end{figure}

Fig.~\ref{fig:star_pattern_come_back_origin_yaw_plots} provides a closer look at the heading control during these trips. By comparing the outbound and inbound data, we see a clear kinematic symmetry. The controller handles the $180^\circ$ turnaround required for the return trip just as effectively as the forward motion, showing similar settling times in both directions. The fact that the robot could consistently retrace its path confirms that the feedback loop effectively handles the ground slip inherent in snake locomotion.

\begin{figure}[t!]
    \centering
    \includegraphics[width=1\linewidth]
    {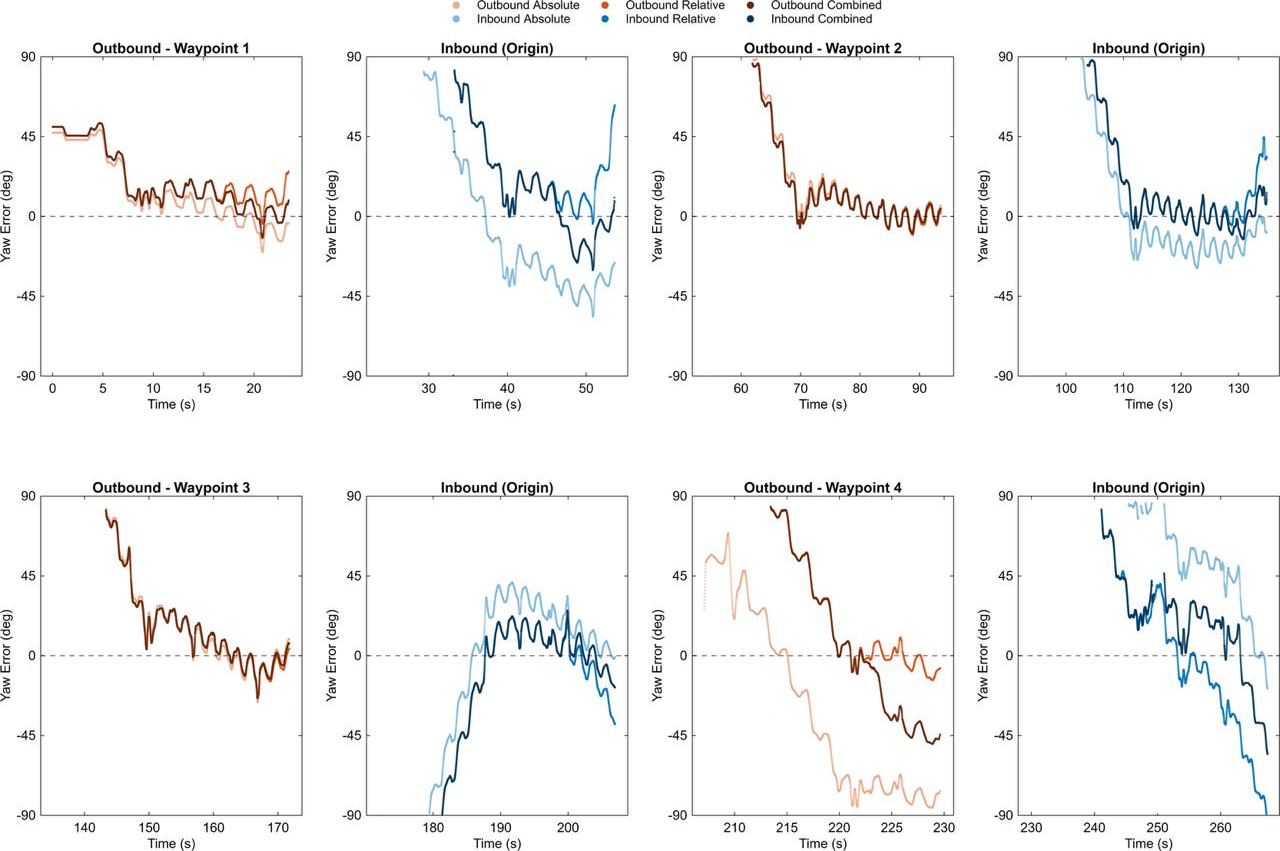}
    \caption{Single-waypoint convergence analysis comprising 13 experiments.}
    \label{fig:star_pattern_come_back_yaw_vs_time}
\end{figure}

\subsection{Evaluation of Robustness to Unmodeled External Disturbances}

The final experiment evaluated the controller’s robustness to unmodeled external disturbances. While traversing a simple path between two waypoints, the robot was intentionally perturbed during locomotion through impulsive pushes that displaced its position and forced rotations that altered its yaw.

\begin{figure}[t!]
    \centering
    \includegraphics[width=1\linewidth]{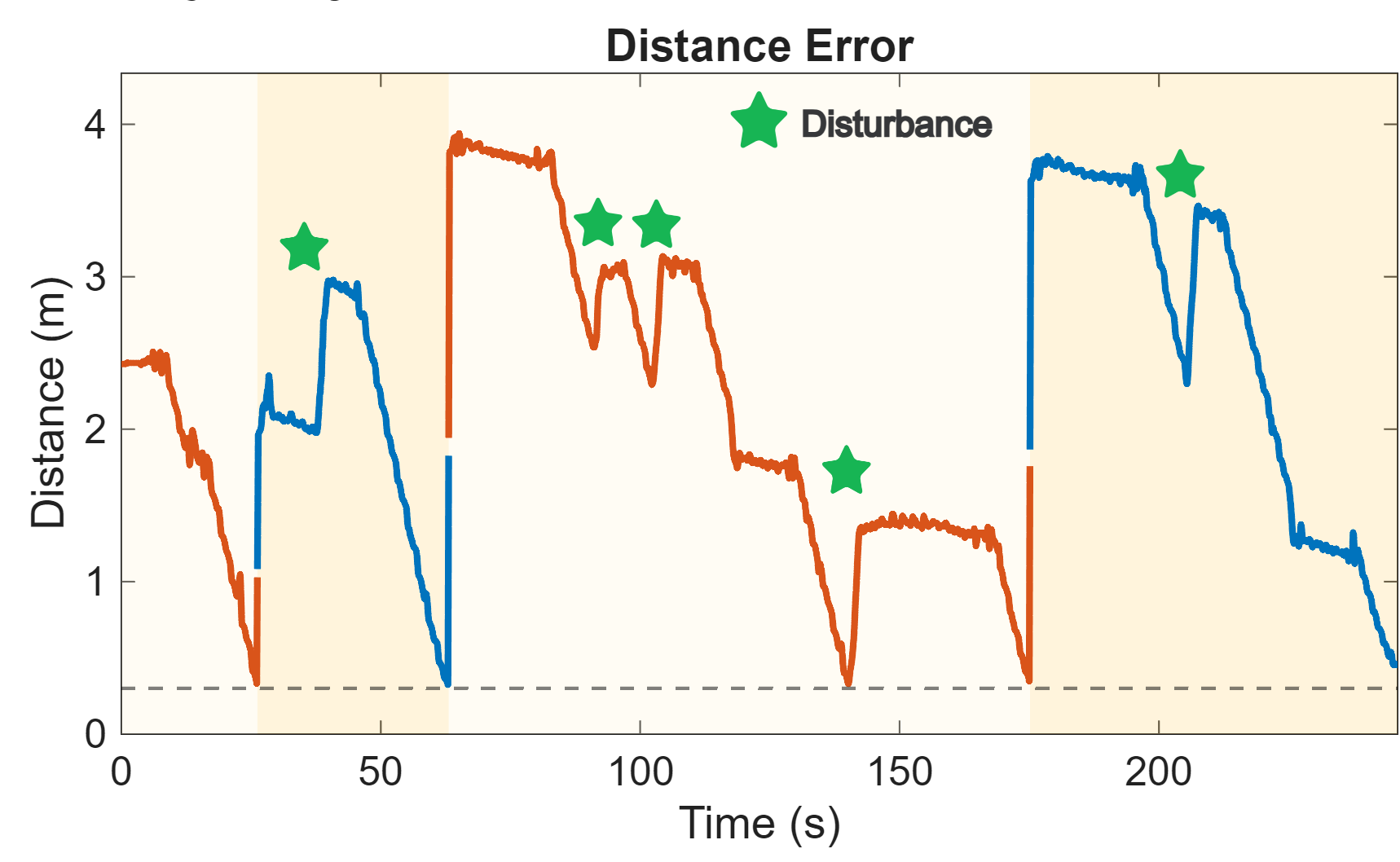}
    \caption{Illustrates the disturbance rejection tracking experiment performance: (a) CoM Trajectory, (b) Distance Error, (c) Yaw Error.}
    \label{fig:star_pattern_come_back_origin_disturb}
\end{figure}

Fig.~\ref{fig:star_pattern_come_back_origin_disturb} summarizes the tracking performance. The closed-loop system consistently rejected these disturbances without requiring any manual reset or intervention. When the robot was physically displaced, the resulting spike in distance error was immediately detected, prompting the controller to adjust the gait and drive the robot back toward the intended path. Similarly, when the robot’s heading was twisted away from the target direction, the controller modulated the turning gait to restore the correct orientation.







 \chapter{Conclusion}
\label{chap:conclusion}

This thesis presented a complete autonomy pipeline for enabling closed-loop waypoint navigation on the COBRA snake robot platform. The work addressed a fundamental gap in serpentine robotics: while snake robots offer exceptional mobility across challenging terrain through their highly articulated bodies and diverse gaits, their autonomous operation in GPS-denied environments has remained largely unexplored. The contributions of this research span perception, state estimation, and control, culminating in a validated system capable of autonomous waypoint tracking without reliance on external infrastructure.

The first major contribution established the feasibility of running computationally intensive visual-inertial SLAM algorithms on compact edge hardware during the highly dynamic and oscillatory motion characteristic of snake robot locomotion. By integrating RTAB-Map with the Intel RealSense D435i depth camera on the NVIDIA Jetson Orin NX, the system achieved real-time pose estimation at 30 Hz while the robot executed sidewinding and vertical undulation gaits. Validation against OptiTrack motion capture ground truth demonstrated the best result having a mean position error of 6.85 cm and RMSE of 7.33 cm, confirming that visual-inertial odometry provides sufficiently accurate state feedback for closed-loop control despite the challenging viewing conditions imposed by the undulating head motion.

The second contribution systematically characterized the visual perceptual challenges unique to snake robot locomotion. Five primary failure modes were identified: low-texture environments causing sparse feature extraction, close proximity to objects exceeding the depth camera's minimum range, sudden jerky motions during gait transitions, head impacts with the ground during vertical undulation, and pure rotational maneuvers lacking sufficient parallax cues. This characterization establishes a baseline understanding of serpentine state estimation limitations and informs the design of robust autonomy systems for future platforms.

The third contribution developed a reduced-order control framework that abstracts the complex 11-DOF kinematic chain into a stable Center-of-Mass reference frame. By computing a mass-weighted average of link positions and constructing a virtual chassis orientation through SVD-based frame computation, the controller operates on a simplified two-dimensional pose representation while the underlying CPG-based locomotion handles the high-dimensional joint coordination. This hierarchical abstraction enables efficient trajectory tracking without requiring full-body dynamic modeling.

The fourth contribution implemented and validated a weighted waypoint-following strategy that dynamically balances relative bearing error for path convergence against absolute orientation error for final pose alignment. The distance-dependent blending function, implemented through smoothstep transitions with trapezoidal weighting, ensures smooth behavioral transitions as the robot approaches each waypoint. Physical experiments validated the complete system across four scenarios: single-waypoint convergence from 13 arbitrary initial configurations, complex multi-waypoint tracking through a star topology, bidirectional trajectory consistency, and disturbance rejection under external perturbations. In all cases, the controller successfully regulated both position and orientation errors, demonstrating robust autonomous navigation capability.

\section{Future Scope}

While this thesis establishes the foundational autonomy pipeline for COBRA, several promising directions remain for future investigation.

The current experimental validation utilized OptiTrack motion capture at 100 Hz for ground-truth pose feedback, but deploying the closed-loop controller with VIO estimates directly presents nontrivial challenges: the VIO operates at 30 Hz compared to OptiTrack's 100 Hz, introducing increased latency and reduced temporal resolution, while accumulated drift will require adaptive control strategies that account for estimation uncertainty or loop closure detection to bound drift during extended missions. 

The controller currently employs fixed CPG parameters optimized for flat, high-friction surfaces, but extending the system to autonomously adjust gait parameters based on perceived terrain properties (through estimating surface friction via proprioceptive feedback, detecting slip events through discrepancies between commanded and observed motion, and modulating wave amplitudes or frequencies accordingly) would significantly expand COBRA's operational envelope. 

Additionally, experimental validation revealed RTAB-Map failures in shadowed regions, requiring active illumination sources synchronized with the camera system for planetary missions targeting permanently shadowed craters.

COBRA's locomotion plasticity encompasses six distinct gait families, yet the current autonomy pipeline addresses only sidewinding; extending the framework to incorporate mode transitions (such as switching from sidewinding to hex-ring tumbling when approaching steep slopes) would leverage the platform's full morphological capabilities through mode selection logic based on terrain assessment and smooth handoffs between gait-specific controllers. 

Furthermore, the current system assumes obstacle-free environments with predefined waypoint sequences, but integrating the dense depth information from the RealSense camera for local obstacle detection, combined with reactive avoidance behaviors or sampling-based path planners, would enable navigation through cluttered environments without complete prior knowledge. These directions build upon the validated perception and control foundations established in this thesis, working toward deploying autonomous snake robots for planetary exploration and terrestrial inspection tasks in unstructured environments.

\bibliographystyle{IEEEtran}  

\bibliography{bib/thesis}

\appendix


\end{document}